\pdfoutput=1

\documentclass[11pt]{article}
\usepackage[table]{xcolor}
\usepackage{acl}
\usepackage{multirow, makecell}
\usepackage{tabularx}
\usepackage{colortbl}
\usepackage{times}
\usepackage{mathrsfs}
\usepackage{arydshln}
\usepackage{latexsym}
\usepackage{booktabs,siunitx,multirow}
\usepackage{amsmath} 
\usepackage[T1]{fontenc}
\usepackage{tabularray}
\usepackage{graphicx} 
\usepackage{subcaption}
\usepackage{booktabs}
\usepackage{todonotes}
\usepackage{nicematrix,tikz}
\usepackage{hyperref}

\usepackage[utf8]{inputenc}
\usepackage{xspace}
\usepackage{microtype}
\usepackage[usestackEOL]{stackengine}
\newcolumntype{P}[1]{>{\centering\arraybackslash}p{#1}}
\newcommand{\splitcell}[1]{\Centerstack{#1}}

\title{Adapting Fake News Detection to the Era of Large Language Models}
\author{Jinyan Su$^1$,   Claire Cardie$^1$,
  Preslav Nakov$^2$\\
  $^1$Department of Computer Science, Cornell University\\
  $^2$Mohamed bin Zayed University of Artificial Intelligence\\
  \texttt{\{js3673,ctc9\}@cornell.edu, preslav.nakov@mbzuai.ac.ae}
  }

\begin{document}
\maketitle
\begin{abstract}
In the age of large language models (LLMs) and the widespread adoption of AI-driven content creation, the landscape of information dissemination has witnessed a paradigm shift. With the proliferation of both human-written and machine-generated real and fake news, robustly and effectively discerning the veracity of news articles has become an intricate challenge. While substantial research has been dedicated to fake news detection, it has either assumed that all news articles are human-written or has abruptly assumed that all machine-generated news was fake. Thus, a significant gap exists in understanding the interplay between machine-paraphrased real news, machine-generated fake news, human-written fake news, and human-written real news. In this paper, we study this gap by conducting a comprehensive evaluation of fake news detectors trained in various scenarios. Our primary objectives revolve around the following pivotal question: How can we adapt fake news detectors to the era of LLMs?
Our experiments reveal an interesting pattern that detectors trained exclusively on human-written articles can indeed perform well at detecting machine-generated fake news, but not vice versa. Moreover, due to the bias of detectors against machine-generated texts \cite{su2023fake}, they should be trained on datasets with a lower machine-generated news ratio than the test set. Building on our findings, we provide a practical strategy for the development of robust fake news detectors. \footnote{The data and the code can be found at \url{https://github.com/mbzuai-nlp/Fakenews-dataset}}
\end{abstract}

\section{Introduction}


Since Brexit and the 2016 US Presidential campaign, the proliferation of fake news has become a major societal concern \cite{martino2020survey}. On the one hand, false information is easier to generate but harder to detect \cite{10.1145/3377330.3377334}. 

\begin{figure}[tbh]
  \centering
  \includegraphics[width=.8\linewidth]{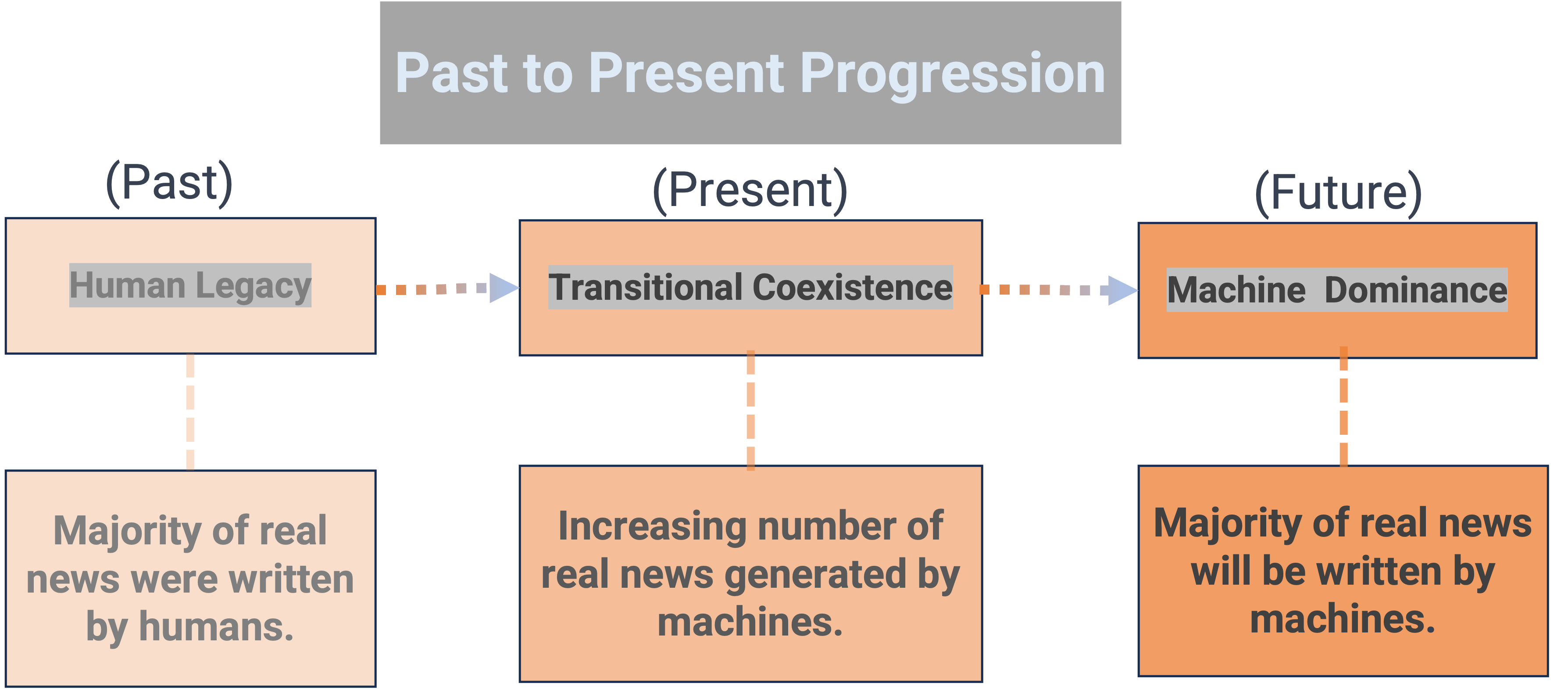}  
\caption{The three phases of transitioning from human-written to machine-generated real news production: (\textit{Human Legacy}, \textit{Transitional Coexistence}, and \textit{Machine Dominance}).}
\label{fig: experimental setting based on real news source}
\end{figure}

On the other hand, people are often attracted to sensational information and studies have shown that it spreads six times faster than truthful news \cite{vosoughi2018spread}, which is a major threat to both individuals and society. 

Until recently, most online disinformation was human-written \cite{vargo2018agenda}, but now a lot of it is AI-generated \cite{simon2023misinformation}. With the progress in LLMs \cite{radford2019language, brown2020language, chowdhery2022palm}, AI-generated content is becoming much harder to detect \cite{wang2024m4gtbench,semeval2024task8,wang-etal-2024:m4}. Moreover, machine-generated text is often perceived as more credible \cite{kreps2022all} and trustworthy \cite{zellers2019defending,spitale2023ai} than human-generated propaganda. This raises pressing concerns about the unprecedented scale of disinformation production that AI models have enabled \cite{bommasani2021opportunities,buchanan2021truth,kreps2022all,augenstein2023factuality,goldstein2023generative,pan-etal-2023-risk,wang2024factuality}. 

While efforts to combat machine-generated fake news date back to as early as 2019 \cite{zellers2019defending}, the majority of research in this field has primarily focused on detecting machine-generated text, rather than evaluating the factual accuracy of machine-generated news articles \cite{huang-etal-2023-faking}. In these studies, machine-generated text is considered to be always fake news, regardless of the factuality of its content.

Previously, when generative AI was less prevalent, it was arguably reasonable to assume that most automatically generated news articles would be primarily used by malicious actors to craft fake news. However, with the remarkable advancement of generative AI in the last two years, and with their introduction in various aspects of our lives, these tools are now broadly adopted for legitimate purposes such as assisting journalists in content creation. Reputable news agencies, for instance, use AI to draft or enhance their articles \cite{hanley2023machine}. Nevertheless, the age-old problem of human-written fake news remains.

This diverse blend of machine-generated genuine news, machine-generated fake articles, human-written fabrications, and human-written factual articles has shifted the way of news generation and the intricate intermingling of content sources is likely to endure in the foreseeable future. 

In order to adapt to the era of LLMs, the next generation of fake news detectors should be able to handle the mixed-content landscape of human/machine-generated real/fake news.
While there exists a substantial body of research on fake news detection, it typically focuses exclusively on human-written fake news \cite{perez2017automatic,khattar2019mvae,kim2018leveraging} or on machine-generated fake news \cite{zellers2019defending,goldstein2023generative,zhou2023synthetic}, essentially framing the problem as detection of machine-generated text. However, robust fake news detectors should primarily assess the authenticity of the news articles, rather than relying on other confounding factors, such as whether the article was machine-generated.
Thus, there is a pressing need to understand fake news detectors on machine-paraphrased real news (\textbf{MR}), machine-generated fake news (\textbf{MF}), human-written fake news (\textbf{HF}), and human-written real news (\textbf{HR}). 

Here, we bridge this gap by evaluating fake news detectors trained with varying proportions of machine-generated and human-written fake news. Our experiments yield the following key insights:

(1) Fake news detectors, when trained exclusively on human-written news articles (i.e., HF and HR), have the ability to detect machine-generated fake news. However, the reverse is not true, i.e., if we train exclusively on machine-generated fake news, the model is worse at detecting human-written fake news. This observation suggests that, when the proportion of testing data is uncertain, it is advisable to train detectors solely on human-written real and fake news articles. Such detectors are still able to generalize effectively for detecting machine-generated news.

(2) Although the overall performance is mainly decided by the distribution of machine-generated and human-written fake news in the test dataset, the class-wise accuracy for our experiments suggests that, 
in order to achieve a balanced performance for all subclasses, we should train the detector on a dataset with a lower proportion of machine-generated news compared to the test set.

(3) Our experiments also reveal that fake news detectors are generally better at detecting machine-generated fake news (MF) than at identifying human-written fake news (HF), even when exclusively trained on human-generated data (without seeing MF during the training). This underscores the inherent bias within fake news detectors \cite{su2023fake}. We recommend to take these biases into consideration when training fake news detectors. 

Our main contributions can be summarized as follows:
\begin{itemize}
\item 

We are the first to conduct comprehensive evaluation of fake news detectors across diverse scenarios where news articles exhibit a wide range of diversity, including both human-written and machine-generated real and fake content.

\item Drawing from our experimental results, we offer  valuable insights and practical guidelines for deploying fake news detectors in real-world contexts, ensuring that they remain effective amid the ever-evolving landscape of news generation.

\item Our work lays the groundwork for understanding the data distribution shifts in fake news caused by LLMs, moving beyond simple fake news detection.

\end{itemize}
\section{Related Work}
Fake news detection is the task of detecting potentially harmful news articles that make some false claims \cite{oshikawa2018survey}. The conventional solution for detecting fake news is to ask professionals such as journalists to perform manual fact-checking \cite{shao2016hoaxy,Survey:2021:AI:Fact-Checkers}, which is expensive and time-consuming. 

To reduce the time and the efforts for detecting fake news, researchers formulated this problem as a classification task and proposed various solutions for automatic fake news detection from a machine learning perspective \cite{baly2018predicting,guo2022survey,10.1145/3517214}. 

There are two main task formulations: one only consider human-written real vs. fake news, and the other one formulates this as detecting machine-generated text, thus automatically categorizing any machine-generated news as fake news. 

\subsection{Human-Written Real vs. Fake News}

Before 2018, fake news was predominantly manually written \cite{vargo2018agenda}, which motivated early research on distinguishing human-written fake news from human-written real news. Various methods have been proposed based on linguistic patterns \cite{rashkin-etal-2017-truth,perez2017automatic}, analysis of the writing style \cite{horne2017just,schuster2020limitations}, and of the content in general \cite{jin2016novel,vv2020fake,vargas2022rhetorical}. Other approaches performed automatic verification of the claims made in news articles \cite{graves2016rise}, analyzed the reliability of the source \cite{baly-etal-2020-written}, or information from social media \cite{FbMultiLingMisinfo}.

\subsection{Distinguishing Machine-Generated from Human-Written News}

With recent progress of natural language text generation \cite{radford2019language}, there have also been rising concerns that malicious actors might generate fake news automatically using controlled generation \cite{zellers2019defending,jawahar2020automatic,huang-etal-2023-faking,mitchell2023detectgpt}. To understand and to respond to neural fake news, \citet{zellers2019defending} studied the potential risk of neural disinformation and presented a model for neural fake news generation called GROVER, which allows for controlled generation of an entire news article. They generated fake news articles using GROVER, and experimented with distinguishing them from real news articles. 
Thus, they essentially addressed the problem of detecting machine-generated vs. human-written news articles, even though they talked about detecting neural fake news. Later work \cite{pagnoni2022threat} discussed different threat scenarios from neural fake news generated by state-of-the-art language models and assessed the performance of the generated-text detection systems under these threat
scenarios. 

Other work proposed more advanced fake news generators that incorporated the use of propaganda techniques \cite{huang-etal-2023-faking}.

With the recent popularity of LLMs, many worry about malicious actors using more powerful models such as ChatGPT, GPT-3, GPT-3.5, and GPT-4 to generate fake news \cite{zhou2023synthetic, hanley2023machine, su-etal-2023-detectllm}.
\citet{pan-etal-2023-risk} studied the risk of misinformation pollution with large language models. \citet{augenstein2023factuality} discussed the factuality challenges in the era of large language models. See also \cite{wang2024factuality} for a recent survey on the factuality of large language models in the year 2024.

There has also been research on detecting machine-generated content \cite{mitchell2023detectgpt,su-etal-2023-detectllm,he2023mgtbench}, including a recent shared task at SemEval-2024 \cite{semeval2024task8}, based on the M4 dataset \cite{wang-etal-2024:m4}.

\section{Methodology}\label{sec: methodology}

As the proportion of human-written vs. machine-generated content shifts, it is crucial to study the impact on a model's proficiency in differentiating between real and fake news.
Here, we consider three distinct experimental setups, each representing different phases for news article generation due to the evolution of LLMs, as shown in Figure~\ref{fig: experimental setting based on real news source}. We experiment with an LLM as the news generator and we consider the news articles to contain only pure text without other modalities, as in previous fake news detection work \cite{zellers2019defending}.

In the initial \textit{Human Legacy} stage, the news was predominantly human-written. In this setting, we only use human-written real news articles as training data for the real news category. Then, in order to see how the proportion of machine-generated fake news in the training data affects the performance of the detector, we  incrementally introduce machine-generated fake news articles, ranging from 0\% to 100\%. This setting mirrors a past era, where humans were the primary producers of real news.

Transitioning to the \textit{Transitional Coexistence} stage, we reflect the current situation where language models collaboratively contribute to real news article generation. To simplify this setting,  our training data in the real news class contain a human-written and a machine-generated part. This setting reflects the growing influence of LLMs in the news landscape.


\begin{table}[tbh]
\centering
\small
\begin{tabular}
{lrrrr}
\toprule
Dataset& HF	&MF	&HR	&MR\\
\midrule
\texttt{GossipCop++}&	4,084&	4,084&	8,168	&4,169\\
\texttt{PolitiFact++}&	97&	97&	194&132\\
\bottomrule
\end{tabular}
\caption{Number of news articles from each subclass in the \texttt{GossipCop++} and \texttt{PolitiFact++} datasets.}
\label{tab: original dataset statistics}
\end{table}

Finally, in the \textit{Machine Dominance} stage, we model a future where machine-generated texts surge for real news generation. For this, the training data for the real news class contains exclusively machine-generated real news articles. This reflects a future where LLMs become the primary and dominant way to produce the news.

\subsection{Data} 

Our data is based on \texttt{GossipCop++} and \texttt{PolitiFact++}, which were introduced in \cite{su2023fake}. They contain human-written fake (\textbf{HF}) and human-written real news (\textbf{HR}) from the FakeNewsNet \cite{shu2018fakenewsnet}, which were filtered to keep only the subset that contains a title and a description. We first sampled 4,084 fake news and 4,084 real news from \texttt{GossipCop++} and then we randomly split these 8,168 examples into 60\% for training, 20\% for validation, and 20\% for testing. For out-of-domain testing, we sampled 97 real and 97 fake news from \texttt{PolitiFact++}. We further generated machine-paraphrased real news (\textbf{MR}) and machine-generated fake news (\textbf{MF}) using ChatGPT and \textit{Structured Mimicry Prompting} \cite{su2023fake} to reduce the identifiable structure of machine-generated news articles, so that the detector can focus on the content rather than on the source. 
Table~\ref{tab: original dataset statistics} shows statistics about our dataset. More analysis and details about the dataset are given in Appendix~\ref{Appendix: dataset}.  

\subsection{Evaluation Measures}
Since we had a balanced training and testing dataset in all our experiments, we use subclass-wise accuracy as our primary evaluation measure. Other measures such as F1, precision, recall, and overall accuracy can be directly derived from the subclass-wise accuracy due to the balanced (sub)class setting. For our purposes, subclass-wise accuracy offers a more direct and insightful perspective, allowing us to assess the results from the standpoint of each individual subclass while considering more measures such as the internal bias of the detector.

\subsection{Experiments}
In our experiments, we used transformer-based methods, as they have demonstrated significantly superior performance compared to other deep learning classifiers and have gained widespread acceptance and adoption in the field of fake news detection \cite{alam-etal-2021-fighting-covid,10.1145/3517214}. In particular, we experimented with both large and base models of BERT \cite{devlin2018bert}, RoBERTa \cite{liu2019roberta}, ELECTRA \cite{clark2020electra}, ALBERT \cite{lan2019albert}, and DeBERTa \cite{he2020deberta}.


\subsection{Experimental Details}

We trained all models on an A100 40G GPU with a batch size of 25 and a learning rate of 1e-6 for 10 epochs.

\section{Experimental Results}
In this section, we describe our exhaustive experiments and exploration of the three stages that we described in Section~\ref{sec: methodology}. Specifically, we evaluate the above-mentioned five transformer-based models of two distinct sizes (base and large) across the three stages. Coupled with the five different proportions of machine-generated fake news, this results in a total of 50 unique model configurations.  We tested each of these configurations on the above-described in-domain test dataset \texttt{GossipCop++} and on the out-of-domain dataset \texttt{PolitiFact++}. 

As we show in Appendix~\ref{Appendix: dataset}, there are sizable differences between \texttt{GossipCop++} and \texttt{PolitiFact++}, and thus the latter can serve as a valuable out-of-domain dataset for assessing the robustness of fake news detectors that were trained on the former.

\subsection{Main Results}
 
\begin{figure*}[tbh]
  \centering
  \includegraphics[width=1\linewidth]{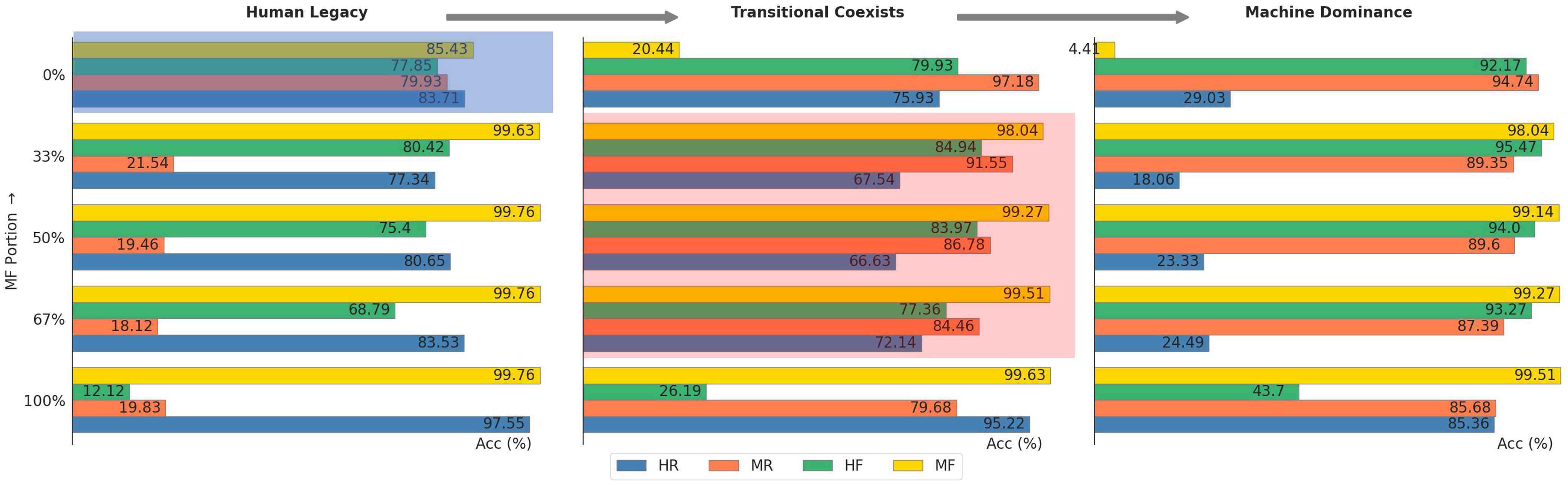}  
\caption{Class-wise detection accuracy from the \textit{Human Legacy} stage (left), to the \textit{Transitional Coexistence} stage (middle), to the \textit{Machine Dominance} stage (right), with different fractions of machine-generated fake news in the fake news training data, shown on the $y$ axis. The blue- and the red-shaded areas are recommended training strategies based on our experiments. We discuss this in detail in Section~\ref{subsec: training strategy}.}
\label{fig: main result}
\end{figure*}

Given the sheer volume of the experiments, to maintain clarity and to avoid overwhelming the readers, we relegate the complete results to Appendix~\ref{Apendix: complete result}, while focusing our analysis and discussion primarily on Figure~\ref{fig: main result}, which shows the performance measures obtained from training RoBERTa-large and testing on the \texttt{GossipCop++} dataset.
 
In order to provide a thorough understanding of our experimental results, we first delve into each stage independently, and then we perform a more holistic analysis of the observed patterns across these stages.

\paragraph{\textit{Human Legacy} Setting.} In this setting, the training data for the real news class is all human-written real news. When paired with human-written fake news as the whole training set, it can achieve a relatively balanced and high detection accuracy for each subclass. 
When the fraction of MF increases to 33\%, the fake news detection accuracy for the MF subclass increases to around 99\%; further increases in the fraction of MF examples in the training data almost has no more contribution to the test detection accuracy for the MF subclass. Moreover, we find an abrupt drop of detection accuracy for the MR subclass.
This might be because, when we add MF examples to the training data, since we do not have any MR examples during training, the detector might use a shortcut such as features that are unique to machine-generated text as features for fake news, and thus could classify most of the MR examples as fake news. Similarly, when the fraction of MF examples increases from 67\% to 100\%, (i.e., we only use machine-generated fake news paired only with human-written real news as training data), we observe an abrupt drop in accuracy for the HF subclass: detectors trained in this way categorize most of the human-written fake news as real, since they check whether the text is machine-generated as a key feature for detecting fake news. Note that, even when the fraction of MF examples is high, the accuracy for the MR subclass is still greater than $1-\text{Acc}(\textbf{MF})$. This suggests that the detector can still learn some features to predict the factuality of the machine-generated texts rather than solely using features for detecting machine-generated text. Otherwise, we would have had $\text{Acc}(\textbf{MR}) \approx 1-\text{Acc}(\textbf{MF})$.

One key observation from this stage is when the proportion of MF is 0\%, which corresponds to a setting where we train a detector on human-written real and fake news articles and we then deploy it to detect machine-generated real and fake news. Interestingly, the resulting detector can generalize well to distinguishing between real and fake machine-generated news, with a detection accuracy almost comparable to detecting human-written ones. This suggests that maybe it is not essential to train on machine-generated real and fake news to be able to detect them. It would certainly be helpful for the overall detection accuracy if our training data distribution aligned well with the testing data; however, in real-world deployment, due to the distribution shift or due to our ignorance about the distribution of the test data (for example, we do not know how many of the news articles are machine-generated, and more importantly, this distribution might change over time due to model updates and other factors \cite{omar2022quantifying}), the most effective way to train the detector is to train on human-written real and fake news articles.

\paragraph{\textit{Transitional Coexistence} Setting.} In this setting, the training data for the real news class is composed equally of machine-generated and human-written articles. Notably, we observe that when the fake news training data is exclusively human-written, the subclass-wise accuracy for the MF subclass is relatively low, with just 20.44\%, while the HF class is accurately detected, with 79.93\% detection accuracy. Conversely, when the fake news class is entirely MF, the accuracy for the HF subclass diminishes to a mere 26.19\%, while the MF accuracy is high. 

Echoing our prior analysis from the \textit{Human Legacy} stage, this may be attributed to the detectors leveraging features that are indicative of an article's source (machine or human) rather than of its veracity. In the absence of HF examples in the training data, the detector may use a shortcut and assume that all fake news are machine-generated, which results in reduced accuracy for the HF subclass. A similar situation arises when no MF data is present during training, potentially leading the detector to misclassify MF articles as real news at test time.

Moreover, even with a balanced fake news class containing half MF and half HF examples, the detection accuracy for the MF subclass consistently surpasses other subclasses, while for HR it is the lowest. This detection accuracy is not as balanced as training on only HF and HR (see the result for the \textit{Human Legacy} stage when there is no MF data, the blue-shaded area). This highlights a key insight: striving for perfect balance within each subclass during training might not yield results as good as training solely on human-written real and fake news. However, since training with the other three subclasses (HR, HF, MF) yields better results than training on human-written real and fake news only, the overall performance might be better (depends on the subclass distribution in the test set).

\paragraph{\textit{Machine Dominance} Setting} In this setting, the entire training data for the real news class comprises MR examples only, with no exposure to HR examples at all during training. When the fake news class has only HF training examples (i.e., no MF), the detector excels in discerning HF and MR, seemingly by identifying the origin (machine or human) of the article rather than modeling its factuality. Given that modeling factuality is inherently more challenging than pinpointing the article's source, this approach compromises the detection accuracy for the MF and the HR subclasses. Remarkably, introducing a modest 33\% of MF articles to the training data triggers a dramatic surge in MF detection accuracy, catapulting it from a mere 4.41\% to an impressive 98.04\%. This swift adaptation suggests, in this training set, that the detector has the capability to discern genuine from counterfeit content without being misled by superficial features classifying MF and MR categories. Such behavior hints at the possibility that the veracity of machine-generated articles (MF and MR) is more discernible than that of human-generated articles (HF and HR). 

\begin{figure}[tbh]
  \centering
  \includegraphics[width=1\linewidth]{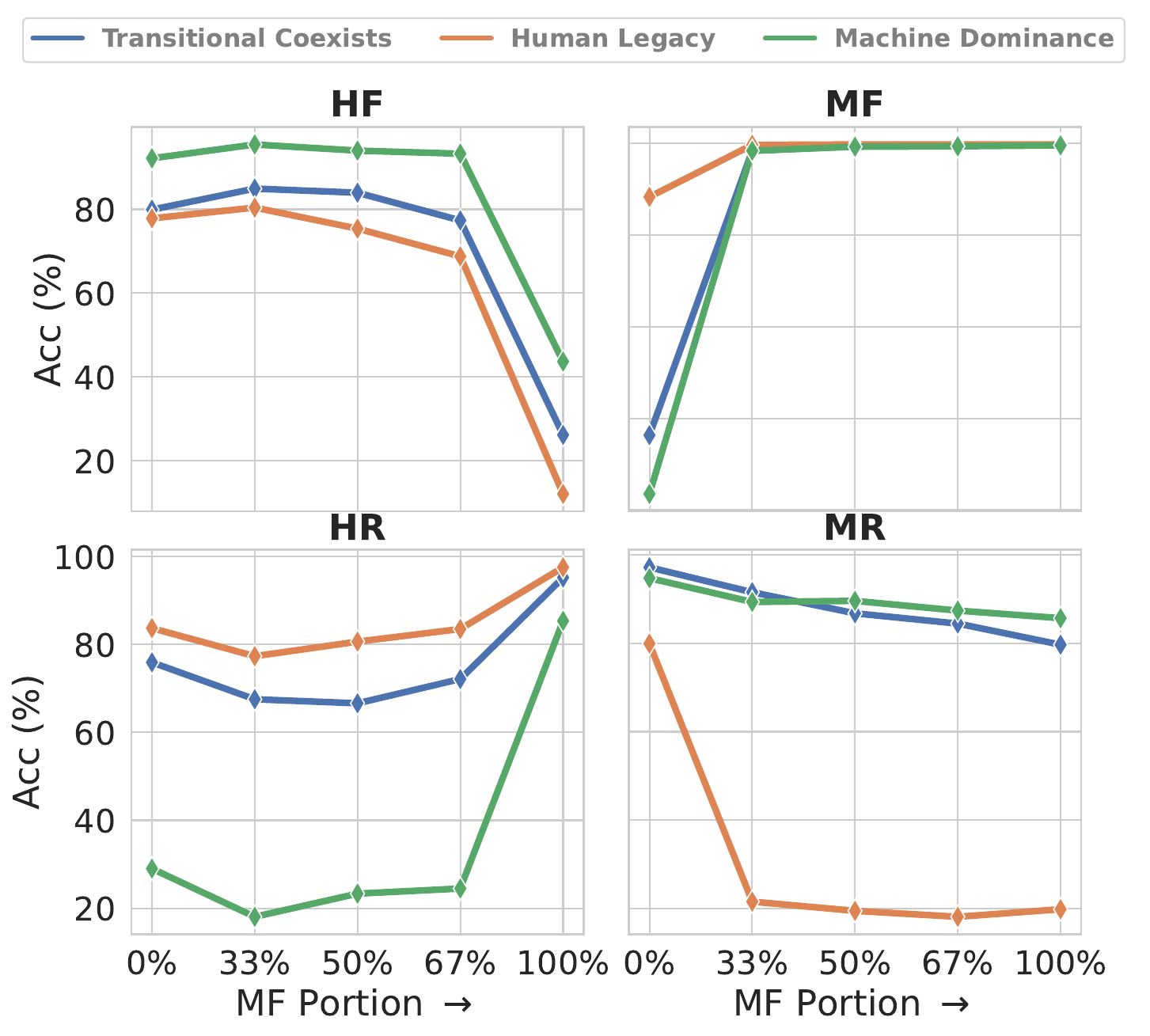}  
\caption{Illustration of the subclass-wise detection accuracy as a function of the fraction of MF examples during training for the three chronological settings.}
\label{fig: class-wise result}
\end{figure}

This hypothesis can be further illuminated by comparing between the \textit{Machine Dominance} setting (with 100\% MF) and the \textit{Human Legacy} one (with 0\% MF), where detectors trained exclusively on human-written articles exhibit commendable accuracy even with machine-generated content, while, in contrast, those trained entirely on machine-generated articles often mistakenly classify the HF subclass as real news.

\subsection{Class-wise Accuracy as a Function of the Fraction of MF Examples in Training}
In this section, we delve into the subclass-wise accuracy for each category. Our primary focus is on understanding how accuracy trends evolve as the proportion of MF examples  increases and discerning the variations in these trends across the different stages. This analysis is illustrated in Figure~\ref{fig: class-wise result}.

\paragraph{Impact of Increasing the Fraction of MF Examples in the Training Data} We can observe in Figure~\ref{fig: class-wise result} some consistent trends across all three stages: as the fraction of MF examples in the training data increases, the accuracy for the MF and the HR subclasses also increases, whereas the accuracy for the HF and the MR subclasses decreases. The improvement for the MF subclass and the decrease for HF are to be expected given that the detectors are exposed to a larger number of MF examples and fewer HF examples during training. The intriguing aspect is the dip in MR detection accuracy and the boost in HR accuracy as the fraction of MF examples increases. 

\begin{figure}[tbh]
  \centering
  \includegraphics[width=1\linewidth]{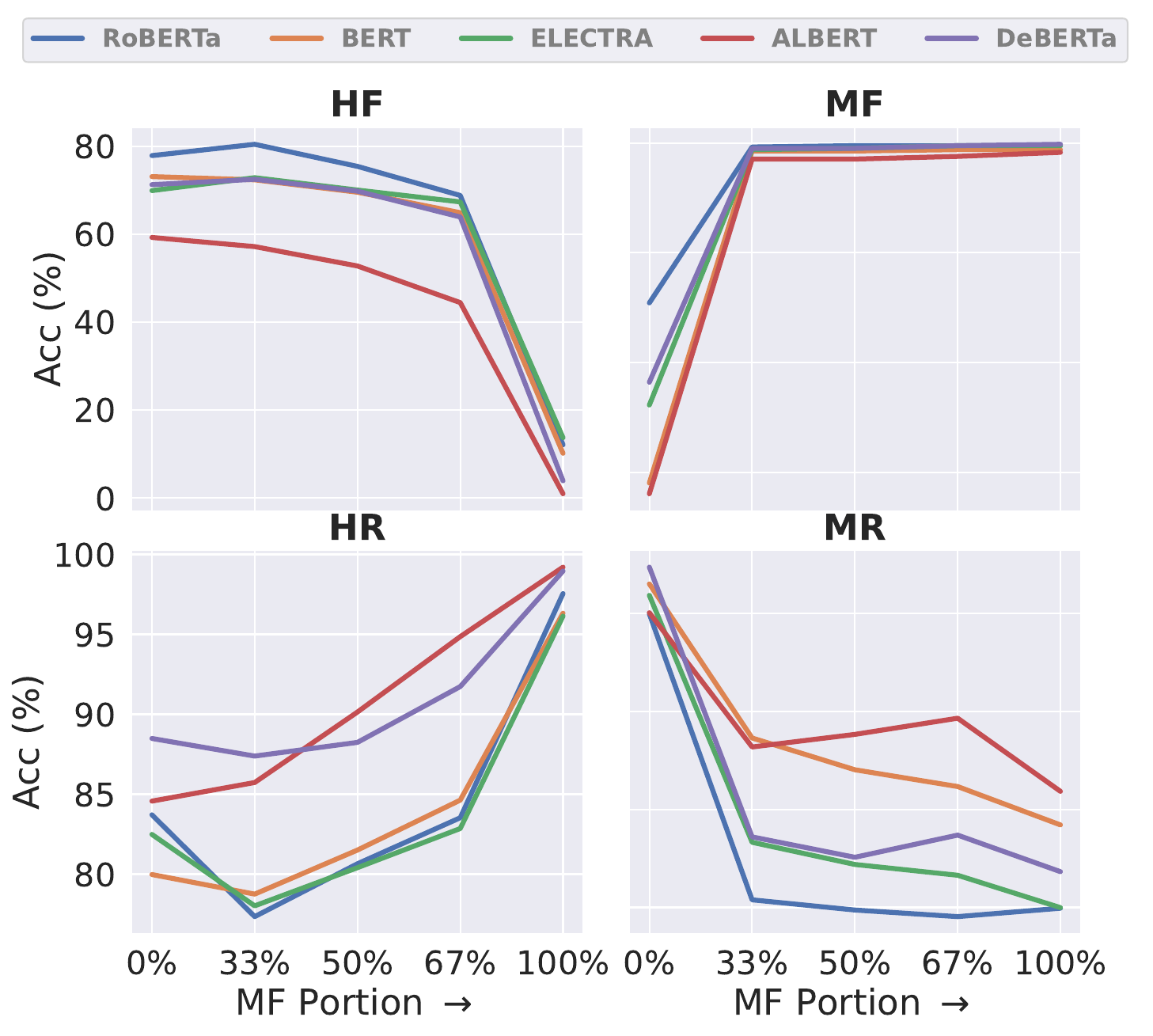}  
\caption{Comparing different detectors (RoBERTa, BERT, ELECTRA, ALBERT, DeBERTa) in the \textit{Human Legacy} setting. }
\label{fig: different detectors}
\end{figure}

Our hypothesis is that, when exposed to more MF training examples, the model increasingly relies on source-related features. Since MR shares confounding features with MF (because they are both machine-generated), their representations are more alike. This similarity might cause the MR examples to be misclassified more frequently as the fraction of MF examples increases. Conversely, the HR subclass, which has the least resemblance to the MF subclass, might get improved accuracy due to the increased presence of MF training examples.

\paragraph{Class-Wise Accuracy Across Stages.}
When examining subclass-wise detection accuracy across stages, the \textit{Transitional Coexistence} setting consistently occupies a median position between the other two stages. Specifically, the \textit{Machine Dominance} setting excels in detecting the HF and the MR subclasses, but it struggles with HR and MF. 

In contrast, in the \textit{Human Legacy} setting the models perform better for the HR and the MF subclasses, but exhibits diminished accuracy for HF and MR. Since the \textit{Machine Dominance} setting predominantly sees machine-generated real news during training, it might become biased towards identifying such patterns, leading to a higher detection rate for HF and MR, but lower for HR and MF. Also, if machine-generated articles have certain consistent patterns, the detector trained predominantly on MR data might rely heavily on them for classification, which affects its performance on HR, which might lack these specific patterns. A similar analysis holds for the \textit{Human Legacy} setting.

\begin{figure}[tbh]
  \centering
  \includegraphics[width=1\linewidth]{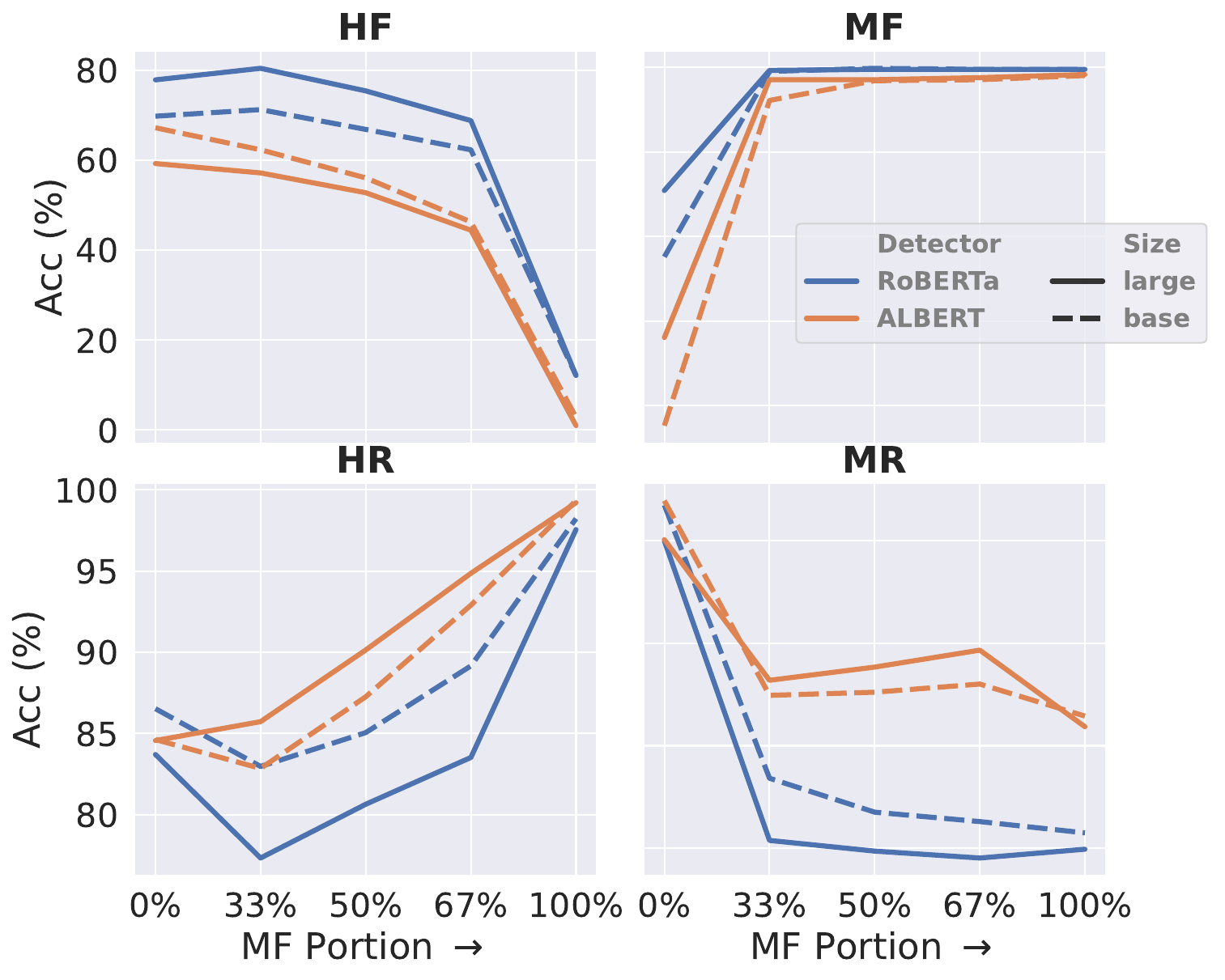}  
\caption{Comparing RoBERTa and ALBERT in the \textit{Human Legacy} setting: large-sized vs. base-sized. }
\label{fig: different detector size}
\end{figure}

\subsection{Analysis of Different Detectors} \label{subsec: detectors}
Below, we compare different detectors: in terms of model architecture and size.

\paragraph{Different Model Architectures.}
In Figure \ref{fig: different detectors}, we compare five detectors: fine-tuned on RoBERTa, BERT, ELECTRA, ALBERT, and DeBERTa (all large-sized models) in the \textit{Human Legacy} setting. We can observe that no model can achieve high detection accuracy for all four subclasses. Instead, there is a trade-off: a detector fine-tuned on RoBERTa achieves the highest detection accuracy for HF and MF, but the lowest accuracy for HR and MR. Meanwhile, a detector fine-tuned on ALBERT achieves the lowest detection accuracy for HF and MF, but the highest accuracy for HR and MR. 

Similar observations can be made about the \textit{Transitional Coexistence} and the \textit{Machine Dominance} settings. You can see more detail in the Appendix~\ref{fig: compare-detectors-MD+Tr}. This might be due to internal model biases: a detector fine-tuned on RoBERTa is more likely to classify articles  as fake news, while such fine-tuned on ALBERT is more likely to classify them as real news.

\paragraph{Impact of Model Size}

To assess how the model size affects detection outcomes, we tested both the large-sized and the base-sized versions of ALBERT and RoBERTa, as shown in Figure~\ref{fig: different detector size}. Interestingly, a larger model does not always outperform the smaller one. In some cases, the smaller model might even mitigate the biases present in the larger variant, yielding better detection results for certain subclasses.

\begin{figure}[tbh]
  \centering
  \includegraphics[width=1\linewidth]{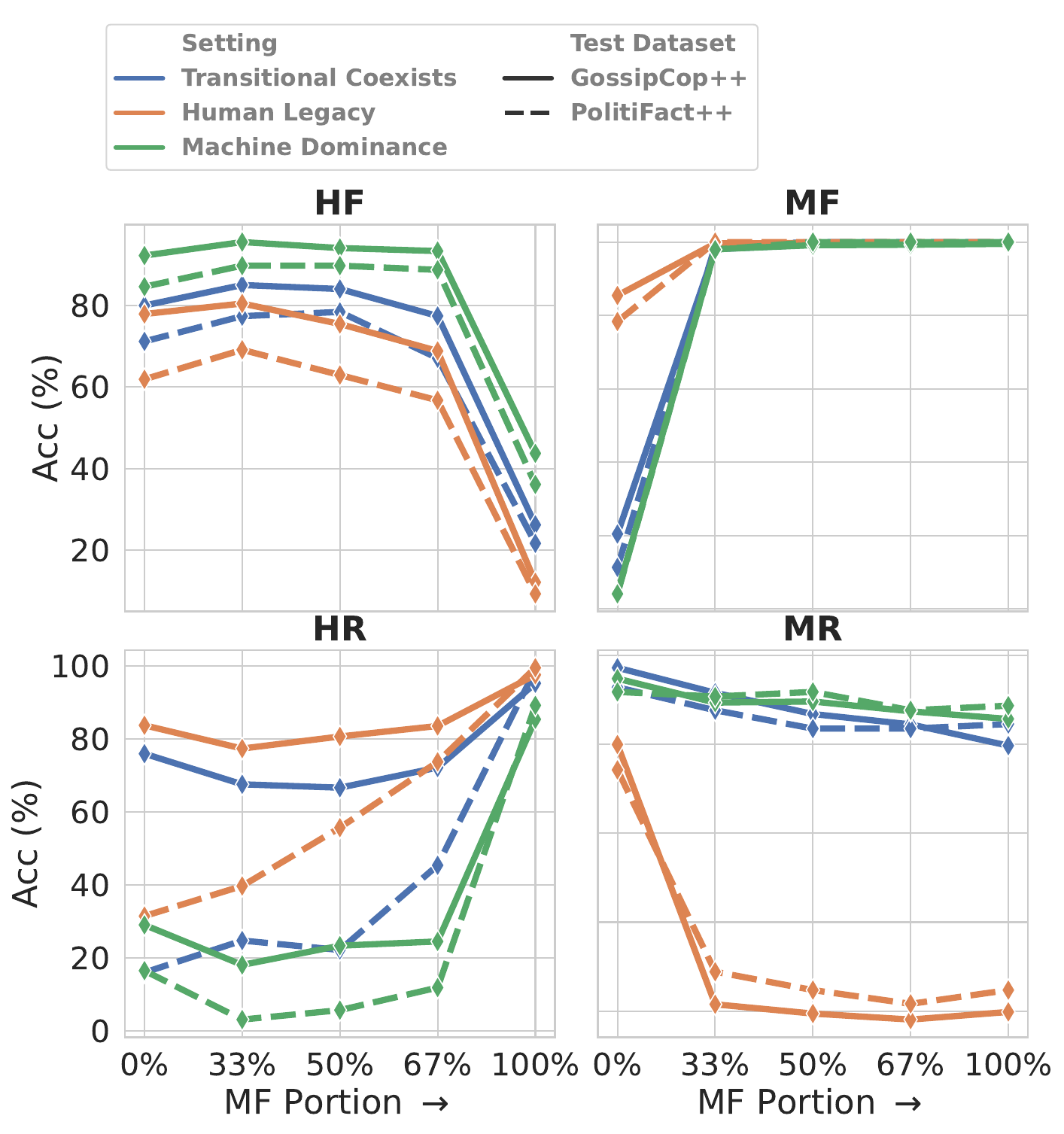}  
\caption{In-domain (\texttt{GossipCop++}) vs. out-of-domain (\texttt{PolitiFact++}) detection. }
\label{fig: out of domain}
\end{figure}

For example, detectors trained on the large-sized ALBERT version show diminished accuracy for the HF subclass compared to the base-sized version. This disparity is even more evident for RoBERTa. Although its larger version adeptly detects HF and MF subclasses, it falters with HR and MR. Conversely, the base-sized RoBERTa model overcomes some of these biases, improving the results for HR and MR, but sacrificing the performance for HF and MF.
Similar trends can also be observed in Figure~\ref{fig: compare-detectors-size-MD+Tr} in the Appendix for the other stages. In summary, no single model size is universally superior. While a larger model might enhance the accuracy for certain subclasses, it might do so at the expense of other subclasses.

\subsection{Out-of-Domain Detection}
In this section, we evaluate the fake news detector on out-of-domain data. The results are shown in Figure~\ref{fig: out of domain}, where lines with the same color are from a similar stage, solid lines are for in-domain, and dashed lines are for out-of-domain testing. We can see that
the detection accuracy declines for almost all subclasses except for MR, where better or equal detection accuracy is achieved when testing on the out-of-domain \texttt{PolitiFact++} dataset. Also, we notice that increasing the proportion of MF examples can help mitigate the gap in the out-of-domain detection accuracy at the expense of the detection accuracy for the HF and the MR subclasses.

\begin{table}[tbh]
\centering
\scriptsize
\resizebox{0.47\textwidth}{!}{
\begin{tabular}
{ccccccc}
\toprule
Subgroup& Training Data& RoBERTa& BERT&ELECTRA&ALBERT& DeBERTa\\
\midrule
\multirow{2}{*}{MR} &All human& \cellcolor[gray]{0.8}
{ -5.7 } & \cellcolor[gray]{0.8}
{ -1.51 } & \cellcolor[gray]{0.8}
{ -3.31 } & \cellcolor[gray]{0.8}
{ -3.88 } & \cellcolor[gray]{0.8}
{ -1.84 } \\
&Mixed& { -3.28 } & { -1.09 } & { 0.58 } & { -2.89 } & { 2.9 } \\\hline
\multirow{2}{*}{MF} &All human& \cellcolor[gray]{0.8}
{ -7.08 } & \cellcolor[gray]{0.8}
{ -8.21 } & \cellcolor[gray]{0.8}
{ -13.25 } & { 8.23 } & \cellcolor[gray]{0.8}
{ -21.51 } \\
&Mixed& { 0.73 } & { 0.21 } & { 1.35 } & \cellcolor[gray]{0.8}
{ 1.33 } & { -0.1 } \\\hline
\multirow{2}{*}{HR} &All human& \cellcolor[gray]{0.8}
{ -52.27 } & \cellcolor[gray]{0.8}
{ -39.77 } & { -7.23 } & \cellcolor[gray]{0.8}
{ -4.67 } & { -30.24 } \\
&Mixed& { -44.46 } & { -39.17 } & \cellcolor[gray]{0.8}
{ -18.43 } & { -0.04 } & \cellcolor[gray]{0.8}
{ -33.68 } \\\hline
\multirow{2}{*}{HF} &All human& \cellcolor[gray]{0.8}
{ -15.99 } & \cellcolor[gray]{0.8}
{ -18.43 } & \cellcolor[gray]{0.8}
{ -22.47 } & { -6.66 } & \cellcolor[gray]{0.8}
{ -16.6 } \\
&Mixed& { -5.62 } & { -11.33 } & { -11.85 } & \cellcolor[gray]{0.8}
{ -23.51 } & { -4.75 } \\
\bottomrule
\end{tabular}}
\caption{Performance degradation in out-of-domain compared to in-domain testing when training on all human data and on mixed data in proportion of HF:MF:HR:MR=1:1:1:1. The gray-shaded part suggests larger performance degradation when evaluated out of domain, and thus less robustness.}
\label{tab: OOD-decrease}
\end{table}

\section{Discussion}
Below, we offer some suggestions about the training data, i.e., how we should balance the machine-generated (MF, MR) and the human-written training data (HF, HR).

\label{subsec: training strategy}

\subsection{In-Domain Detection} 

In the in-domain setting, we found that training with either all human-written data (see the left subfigure of Figure~\ref{fig: main result}, where we highlighted with blue shades) or with a mixture of all four subclasses (see the middle subfigure in Figure~\ref{fig: main result}, which are highlighted with red shades) can achieve a relatively satisfying detection result for all subclasses. 

However, detectors trained with all human-written data (the blue-shaded part) seem to be a better option since it is more balanced on each subclass, while detectors trained on mixtures of all subclasses (the red shaded area) sacrifice HR accuracy for higher MF detection accuracy. Thus, we recommend using only human-written real and fake new articles for training an in-domain detector. 

\subsection{Out-of-Domain Detection} Figure~\ref{fig: out of domain} shows that when increasing the number of MF examples, the margin between in-domain and out-of-domain accuracy decreases. We further calculated the difference between in-domain and out-of-domain accuracy (namely, the class-wise accuracy for \texttt{PolitiFact++} minus the class-wise accuracy for \texttt{GossipCop++}), when trained with only human-written news articles as well as when trained with mixed sources (HF:MF:HR:MR=1:1:1:1). 
The results are shown in Table~\ref{tab: OOD-decrease}. We can see that using mixed training data yields a smaller gap in accuracy. Thus, we recommend to train a detector by adding some MR and MF data to improve the detectors' generalization ability on different domains.

\section{Limitations}
One limitation of our work is that we used a coarse-grained proportion of machine-generated articles for training. Our objective was to offer insights and to highlight potential adaptations in the training strategies during in the age of LLMs, thus raising awareness of responsible use of LLMs, and the three stages we outlined. Note that it is easy to extend our framework to a more fine-grained study.  

The limitation in our paper as well as the observation from the experiments evoke several interesting future directions to address. From the perspective of fake news detection and misinformation research, there is a need for more nuanced evaluation and for combining different detectors to improve the detection accuracy for better fake news detection. Moreover,
our experiments inspire us to generalize the study of real/fake news distribution drift trends to macro contexts, particularly in light of how LLMs influence data distribution shifts. We elaborate more on this below.

\paragraph{More Fine-Grained Evaluation Setting.} 
Our experiments revealed that while training exclusively on human-generated data yields balanced and high accuracy for each subclass relative to the mixed training approach, its robustness is limited for out-of-domain detection. Incorporating some machine-generated data appears to enhance this robustness without significant performance trade-offs. Our current study focused only on the MR proportions of 0\%, 50\%, and 100\%. Further, nuanced experiments are required to pinpoint the optimal balance between class-specific detection accuracy and robustness. It is particularly pertinent to explore MR proportions under 50\% to better assess performance and robustness.
\paragraph{Human-AI co-authorship} In reality, mixed authorship where the text is human-written, but enhanced by a machine, or written by a machine (based on a human prompt) but edited by a human are more likely to be the case. Instead of purely machine-generated or human-written, the above co-authorship is an interesting venue to explore. 

\paragraph{Data Distribution Shift and its Consequences.} Our work delineates three temporal settings: \textit{Human Legacy}, \textit{Transitional Coexistence}, and \textit{Machine Dominance}. These stages offer a simplified view of potential LLM-induced distribution changes, when observed in a longer time span. 

One angle to approach this data distribution shift is via performative prediction \cite{perdomo2020performative}, suggesting that model outputs reciprocally influence data distribution. While there is still a discernible gap between human-written and machine-generated text distributions, the pervasive use of large language models and their outputs might influence the human-written text distribution, and over time, the relative proportion of machine-generated and human-written texts would get closer to each other and might converge to a static landscape. For example, in Figure~\ref{fig: dataset distribution pairwise comparison-gossipcop}, we can observe a distinctive discrepancy for MR and MF, while HF and HR are quite similar. We conjecture that the distribution of the four subclasses might evolve to convergence given a sufficient time horizon. Thus, it would be interesting to analyze fake news detection within an evolving framework.

\paragraph{More Comprehensive Dataset} Since dataset design is not the main focus of the paper, the dataset used might not be comprehensive enough to draw definite conclusions. Thus, a separate work that focuses entirely on the dataset is considered as an interesting and important future research direction. We expect the new dataset to contain multiple fake news generators, multiple languages, and multiple news domains. Moreover, it would be more interesting to contain some side information such as network structures. Note that it is easier to collect such a dataset in the near future than now as LLMs becomes more and more commonly used by news producers.

\section{Ethics and Broader Impact}
Our research delves into fake news detectors and the dynamics of mis/disinformation, positing three hypothetical scenarios. While these scenarios are grounded in reason, they primarily serve to gauge detector performance and behavior. They should not be construed as predictions of the future landscape of fake and real news generation.
Our aim is to raise awareness of the potential risks that LLMs can pose, which goes beyond mis/disinformation and fake news detection, but to more subtle ways of influence related to the proportion of human-written texts online. We thus advocate for a responsible use of LLMs.

\bibliography{anthology,custom}
\bibliographystyle{acl_natbib}
\appendix
\onecolumn
\newpage

\section{Complete Results}\label{Apendix: complete result}
The complete results for the three stages evaluated in our paper are shown in the tables below: for the \textit{Human Legacy} setting in Table~\ref{tab: complete result: All HR}, for the \textit{Transitional Coexistence} setting in Table~\ref{tab: complete result: half MR half HR}, and for the \textit{Machine Dominance} setting in Table~\ref{tab: complete result: All MR}. We show results when using different detectors for in-domain (\texttt{GossipCop++}) and out-of-domain (\texttt{PolitiFact++}) experiments.

\begin{table*}[h]
\centering
\scriptsize
\begin{tabular}
{ccccccc|cccc}
\toprule
&&& \multicolumn{4}{c}{\texttt{GossipCop++}}&\multicolumn{4}{c}{\texttt{PolitiFact++}} \\ \cline{4-11}
&&&\multicolumn{4}{c}{Accurancy w.r.t. each group}&\multicolumn{4}{c}{Accurancy w.r.t. each group} \\ \cline{4-11}
\multirow{2}{*}{{\splitcell{MF portion\\(Training Data)}} 
}&&  &\multicolumn{2}{c}{Real}&\multicolumn{2}{|c}{Fake}&\multicolumn{2}{c}{Real}&\multicolumn{2}{|c}{Fake}  \\ \cline{4-11}
 & Model size & Model name & HR & MR & HF & MF & HR & MR & HF & MF\\ \hline
\multirow{10}{*}{0\%}&\multirow{5}{*}{Large}&RoBERTa & 83.71 & 79.93 & \textbf{77.85} & \textbf{85.43} & 31.44 & 74.23 & 61.86 & 78.35 \\
&&BERT & 79.98 & 86.05 & 73.07 & 69.03 & 40.21 & 84.54 & 54.64 & 60.82 \\
&&ELECTRA & 82.49 & 83.72 & 69.89 & 76.13 & 75.26 & 80.41 & 47.42 & 62.89 \\
&&ALBERT & 84.57 & 80.17 & 59.24 & 68.05 & 79.90 & 76.29 & 52.58 & 76.29 \\
&&DeBERTa & \textbf{88.49} & \textbf{89.47} & 71.24 & 78.21 & 58.25 & 87.63 & 54.64 & 56.70 \\ \cdashline{2-11}
&\multirow{5}{*}{Base}&RoBERTa & 86.53 & 86.90 & 69.77 & 77.60 & 77.84 & 84.54 & 37.11 & 61.86 \\
&&BERT & 86.28 & 84.33 & 63.16 & 78.70 & 76.80 & 85.57 & 30.93 & 69.07 \\
&&ELECTRA & 86.83 & 82.86 & 63.53 & 80.66 & \textbf{90.72} & 80.41 & 40.21 & \textbf{79.38} \\
&&ALBERT & 84.63 & 87.76 & 67.20 & 57.65 & 65.46 & \textbf{88.66} & 57.73 & 56.70 \\
&&DeBERTa & 80.47 & 81.52 & 70.13 & 78.09 & 70.10 & 79.38 & \textbf{74.23} & 78.35 \\
\hline
\multirow{10}{*}{33\%}&\multirow{5}{*}{Large}&RoBERTa & 77.34 & 21.54 & \textbf{80.42} & \textbf{99.63} & 39.69 & 28.87 & \textbf{69.07} & \textbf{100.00} \\
&&BERT & 78.75 & \textbf{54.59} & 72.34 & 99.27 & 44.33 & 50.52 & 60.82 & 97.94 \\
&&ELECTRA & 78.02 & 33.29 & 72.83 & 99.39 & 72.68 & 31.96 & 59.79 & 98.97 \\
&&ALBERT & 85.73 & 52.75 & 57.16 & 98.53 & 81.96 & \textbf{51.55} & 31.96 & 97.94 \\
&&DeBERTa & \textbf{87.39} & 34.39 & 72.46 & 99.51 & 72.16 & 42.27 & 64.95 &\textbf{100.00} \\\cdashline{2-11}
&\multirow{5}{*}{Base}&RoBERTa & 82.98 & 33.66 & 71.24 & 99.51 & 73.71 & 25.77 & 50.52 &\textbf{100.00} \\
&&BERT & 83.71 & 46.14 & 65.97 & 99.39 & 64.95 & 47.42 & 36.08 &\textbf{100.00} \\
&&ELECTRA & 83.28 & 37.33 & 63.04 & 97.92 & \textbf{89.69} & 35.05 & 48.45 &\textbf{100.00} \\
&&ALBERT & 82.85 & 49.82 & 62.30 & 96.08 & 71.13 & 50.52 & 40.21 & 97.94 \\
&&DeBERTa & 87.08 & 39.29 & 64.63 & 98.65 & 81.96 & 36.08 & 62.89 & 98.97 \\
\hline
\multirow{10}{*}{ 50\%}&\multirow{5}{*}{Large}&RoBERTa & 80.65 & 19.46 & \textbf{75.40} & 99.76 & 55.67 & 24.74 & \textbf{62.89} & \textbf{100.00} \\
&&BERT & 81.51 & 48.10 & 69.52 & 99.27 & 45.88 & 46.39 & 51.55 & 97.94 \\
&&ELECTRA & 80.40 & 28.76 & 70.01 & 99.51 & 82.99 & 27.84 & 52.58 &\textbf{100.00} \\
&&ALBERT & \textbf{90.14} & \textbf{55.32} & 52.75 & 98.53 & \textbf{91.75} & \textbf{53.61} & 27.84 & 98.97 \\
&&DeBERTa & 88.24 & 30.23 & 69.77 & 99.51 & 64.95 & 34.02 & 57.73 &\textbf{100.00} \\\cdashline{2-11}
&\multirow{5}{*}{Base}&RoBERTa & 85.06 & 27.05 & 66.83 & \textbf{99.88} & 83.51 & 23.71 & 40.21 &\textbf{100.00} \\
&&BERT & 85.73 & 44.68 & 62.67 & 99.39 & 70.10 & 46.39 & 34.02 &\textbf{100.00} \\
&&ELECTRA & 85.55 & 33.41 & 61.32 & 99.27 & 91.24 & 30.93 & 42.27 &\textbf{100.00} \\
&&ALBERT & 87.26 & 50.43 & 56.06 & 98.41 & 81.96 & 51.55 & 31.96 &\textbf{100.00} \\
&&DeBERTa & 89.83 & 35.74 & 59.61 & 99.27 & 90.21 & 32.99 & 47.42 &\textbf{100.00} \\
\hline
\multirow{10}{*}{67\%}&\multirow{5}{*}{Large}&RoBERTa & 83.53 & 18.12 & \textbf{68.79} & \textbf{99.76} & 73.71 & 21.65 & \textbf{56.70} & \textbf{100.00} \\
&&BERT & 84.63 & 44.68 & 64.87 & 99.39 & 60.31 & 39.18 & 40.21 & 97.94 \\
&&ELECTRA & 82.85 & 26.56 & 67.32 & 99.76 & 88.66 & 26.80 & 45.36 &\textbf{100.00} \\
&&ALBERT & \textbf{94.86} & \textbf{58.63} & 44.43 & 98.78 & 96.91 & \textbf{59.79} & 20.62 & 98.97 \\
&&DeBERTa & 91.73 & 34.76 & 63.89 & 99.76 & 75.26 & 38.14 & 47.42 &\textbf{100.00} \\ \cdashline{2-11}
&\multirow{5}{*}{Base}&RoBERTa & 89.16 & 25.21 & 62.30 & 99.76 & 90.21 & 23.71 & 29.90 &\textbf{100.00} \\
&&BERT & 87.75 & 44.31 & 55.20 & 99.51 & 78.35 & 45.36 & 26.80 &\textbf{100.00} \\
&&ELECTRA & 88.36 & 34.27 & 57.65 & 99.39 & 94.85 & 32.99 & 30.93 &\textbf{100.00} \\
&&ALBERT & 92.90 & 52.02 & 46.27 & 98.53 & 92.27 & 52.58 & 20.62 &\textbf{100.00} \\
&&DeBERTa & 92.77 & 29.99 & 47.37 & 99.39 & \textbf{97.42} & 28.87 & 35.05 &\textbf{100.00} \\
\hline
\multirow{10}{*}{100\%}&\multirow{5}{*}{Large}&RoBERTa & 97.55 & 19.83 & 12.12 & 99.76 & 99.48 & 24.74 & 9.28 & \textbf{100.00} \\
&&BERT & 96.33 & 36.84 & 10.16 & 99.39 & 87.63 & 34.02 & \textbf{12.37} &\textbf{100.00} \\
&&ELECTRA & 96.14 & 19.95 & 13.71 & 99.76 & 99.48 & 25.77 & 6.19 &\textbf{100.00} \\
&&ALBERT & 99.20 & 43.70 & 0.98 & 99.14 & 98.97 & \textbf{49.48} & 1.03 & 98.97 \\
&&DeBERTa & 98.96 & 27.29 & 3.92 & \textbf{99.88} & 99.48 & 34.02 & 9.28 &\textbf{100.00} \\ \cdashline{2-11}
&\multirow{5}{*}{Base}&RoBERTa & 98.22 & 23.01 & 12.12 & 99.76 & 98.97 & 25.77 & 3.09 &\textbf{100.00} \\
&&BERT & 98.16 & 41.74 & 6.61 & 99.76 & 96.39 & 43.30 & 4.12 &\textbf{100.00} \\
&&ELECTRA & 94.67 & 28.52 & \textbf{18.97} & 99.76 & 97.42 & 28.87 & 8.25 &\textbf{100.00} \\
&&ALBERT & \textbf{99.33} & \textbf{45.78} & 2.82 & 99.02 & \textbf{100.00} & 48.45 & 4.12 &\textbf{100.00} \\
&&DeBERTa & 98.53 & 28.03 & 7.83 & 99.76 &\textbf{100.00} & 32.99 & 8.25 &\textbf{100.00} \\
\bottomrule
\end{tabular}
\caption{Complete results for the \textit{Human Legacy} setting. }
\label{tab: complete result: All HR}
\end{table*}
\begin{table*}[h!]
\centering
\scriptsize
\begin{tabular}
{ccccccc|cccc}
\toprule
&&& \multicolumn{4}{c}{\texttt{GossipCop++}}&\multicolumn{4}{c}{\texttt{PolitiFact++}} \\ \cline{4-11}
&&&\multicolumn{4}{c}{Accurancy w.r.t. each group}&\multicolumn{4}{c}{Accurancy w.r.t. each group} \\ \cline{4-11}
\multirow{2}{*}{{\splitcell{MF portion\\(Training Data)}} 
}&&  &\multicolumn{2}{c}{Real}&\multicolumn{2}{|c}{Fake}&\multicolumn{2}{c}{Real}&\multicolumn{2}{|c}{Fake}  \\ \cline{4-11}& Model size & Model name & HR & MR & HF & MF & HR & MR & HF & MF\\ \hline
\multirow{10}{*}{0\%}&\multirow{5}{*}{Large}&RoBERTa & 75.93 & 97.18 & \textbf{79.93} & 20.44 & 15.98 & 92.78 & 71.13 & 11.34 \\
&&BERT & 78.08 & 97.43 & 74.30 & 14.32 & 36.60 & \textbf{97.94} & 60.82 & \textbf{15.46} \\
&&ELECTRA & 81.38 & 97.31 & 72.34 & \textbf{27.29} & 30.93 & 94.85 & 68.04 & 6.19 \\
&&ALBERT & 65.52 & 92.53 & 73.68 & 13.34 & 51.55 & 90.72 & 73.20 & 15.46 \\
&&DeBERTa & 75.81 & 96.33 & 77.23 & 24.72 & 39.69 & 91.75 & 61.86 & 4.12 \\\cdashline{2-11}
&\multirow{5}{*}{Base}&RoBERTa & 79.79 & 97.67 & 73.19 & 25.34 & 68.04 & 96.91 & 51.55 & 13.40 \\
&&BERT & 78.02 & 96.94 & 68.67 & 18.85 & 65.98 & 95.88 & 59.79 & 7.22 \\
&&ELECTRA & \textbf{84.75} & \textbf{98.04} & 66.10 & 19.09 & \textbf{84.54} & 95.88 & 46.39 & 1.03 \\
&&ALBERT & 66.69 & 94.61 & 74.66 & 17.01 & 36.60 & 93.81 & 73.20 & 9.28 \\
&&DeBERTa & 63.99 & 94.61 & 79.07 & 18.36 & 40.72 & 89.69 & \textbf{78.35} & 7.22 \\
\hline
\multirow{10}{*}{33\%}&\multirow{5}{*}{Large}&RoBERTa & 67.54 & 91.55 & \textbf{84.94} & \textbf{98.04} & 24.74 & 87.63 & 77.32 & \textbf{98.97} \\
&&BERT & 62.46 & 86.66 & 82.99 & 95.35 & 18.04 & 84.54 & 72.16 & 95.88 \\
&&ELECTRA & 70.73 & 91.19 & 79.19 & 96.33 & 40.72 & 87.63 & 68.04 & 97.94 \\
&&ALBERT & 69.38 & 89.84 & 68.05 & 91.06 & 66.49 & 84.54 & 53.61 & 91.75 \\
&&DeBERTa & 69.63 & 93.76 & 80.29 & 97.06 & 47.42 & \textbf{92.78} & \textbf{81.44} & 95.88 \\\cdashline{2-11}
&\multirow{5}{*}{Base}&RoBERTa & 70.12 & 89.84 & 79.93 & 93.15 & 50.52 & 89.69 & 56.70 & 88.66 \\
&&BERT & 74.59 & 92.04 & 74.05 & 95.47 & 41.75 & 91.75 & 63.92 & 98.97 \\
&&ELECTRA & 72.99 & 89.84 & 72.58 & 88.37 & \textbf{78.87} & 87.63 & 68.04 & 91.75 \\
&&ALBERT & 72.32 & 92.53 & 72.46 & 89.60 & 44.33 & 90.72 & 72.16 & 95.88 \\
&&DeBERTa & \textbf{74.83} & \textbf{94.12} & 73.68 & 91.19 & 48.97 & 87.63 & 80.41 & 88.66 \\
\hline
\multirow{10}{*}{50\%}&\multirow{5}{*}{Large}&RoBERTa & 66.63 & 86.78 & \textbf{83.97} & \textbf{99.27} & 22.16 & 83.51 & \textbf{78.35} & \textbf{100.00} \\
&&BERT & 71.65 & 86.66 & 78.34 & 96.70 & 32.47 & 85.57 & 67.01 & 96.91 \\
&&ELECTRA & 71.52 & 89.11 & 75.76 & 98.65 & 53.09 & 89.69 & 63.92 &\textbf{100.00} \\
&&ALBERT & \textbf{79.42} & 91.55 & 57.53 & 93.51 & 79.38 & 88.66 & 34.02 & 94.85 \\
&&DeBERTa & 76.97 & \textbf{94.00} & 75.89 & 98.04 & 43.30 & \textbf{96.91} & 71.13 & 97.94 \\\cdashline{2-11}
&\multirow{5}{*}{Base}&RoBERTa & 74.89 & 88.13 & 77.23 & 95.84 & 55.67 & 83.51 & 54.64 & 92.78 \\
&&BERT & 78.44 & 90.82 & 70.50 & 96.82 & 54.64 & 91.75 & 55.67 & 98.97 \\
&&ELECTRA & 77.83 & 87.39 & 67.32 & 93.88 & \textbf{85.57} & 90.72 & 58.76 & 94.85 \\
&&ALBERT & 78.81 & 91.06 & 64.38 & 91.92 & 68.04 & 88.66 & 45.36 & 95.88 \\
&&DeBERTa & 76.67 & 92.41 & 70.13 & 94.74 & 66.49 & 85.57 & 77.32 & 94.85 \\
\hline
\multirow{10}{*}{67\% }&\multirow{5}{*}{Large}&RoBERTa & 72.14 & 84.46 & \textbf{77.36} & \textbf{99.51} & 45.36 & 83.51 & \textbf{67.01} & \textbf{100.00} \\
&&BERT & 76.06 & 84.70 & 72.71 & 98.65 & 39.18 & 83.51 & 60.82 & 97.94 \\
&&ELECTRA & 74.65 & 88.74 & 71.60 & 99.39 & 77.32 & 89.69 & 53.61 &\textbf{100.00} \\
&&ALBERT & \textbf{87.32} & 92.41 & 45.90 & 95.47 & 88.66 & 92.78 & 17.53 & 94.85 \\
&&DeBERTa & 84.63 & \textbf{95.10} & 65.97 & 99.14 & 77.32 & \textbf{94.85} & 58.76 &\textbf{100.00} \\ \cdashline{2-11}
&\multirow{5}{*}{Base}&RoBERTa & 76.55 & 84.82 & 73.56 & 98.90 & 75.26 & 82.47 & 40.21 & 98.97 \\
&&BERT & 84.38 & 90.21 & 63.16 & 97.80 & 72.68 & 90.72 & 37.11 & 98.97 \\
&&ELECTRA & 81.14 & 86.78 & 62.30 & 96.45 & 88.14 & 88.66 & 46.39 & 98.97 \\
&&ALBERT & 86.65 & 92.17 & 54.10 & 95.10 & 80.93 & 91.75 & 35.05 & 94.85 \\
&&DeBERTa & 85.06 & 89.23 & 53.12 & 95.96 & \textbf{92.27} & 88.66 & 44.33 & 97.94 \\
\hline
\multirow{10}{*}{100\%}&\multirow{5}{*}{Large}&RoBERTa & 95.22 & 79.68 & \textbf{26.19} & \textbf{99.63} & 98.97 & 84.54 & \textbf{21.65} & \textbf{100.00} \\
&&BERT & 96.02 & 83.48 & 14.81 & 98.41 & 84.02 & 80.41 & 17.53 & 98.97 \\
&&ELECTRA & 95.71 & 86.17 & 21.54 & 99.63 & 96.91 & 84.54 & 16.49 &\textbf{100.00} \\
&&ALBERT & \textbf{99.27} & \textbf{96.08} & 1.96 & 96.57 & \textbf{99.48} & \textbf{97.94} & 2.06 & 95.88 \\
&&DeBERTa & 98.53 & 93.88 & 9.18 & 99.39 & 99.48 & 93.81 & 18.56 &\textbf{100.00} \\ \cdashline{2-11}
&\multirow{5}{*}{Base}&RoBERTa & 95.41 & 78.09 & 24.24 & 99.63 & 97.42 & 76.29 & 6.19 &\textbf{100.00} \\
&&BERT & 96.39 & 86.05 & 9.91 & 98.41 & 90.21 & 85.57 & 11.34 &\textbf{100.00} \\
&&ELECTRA & 93.75 & 85.31 & 25.21 & 98.29 & 95.88 & 85.57 & 16.49 &\textbf{100.00} \\
&&ALBERT & 98.53 & 95.72 & 5.14 & 96.70 & 97.42 & 96.91 & 3.09 & 96.91 \\
&&DeBERTa & 97.80 & 92.41 & 11.75 & 98.90 & 98.45 & 92.78 & 12.37 & 98.97 \\
\bottomrule
\end{tabular}
\caption{Complete results for the \textit{Transitional Coexistence} setting. }
\label{tab: complete result: half MR half HR}
\end{table*}

\begin{table*}[h]
\centering
\scriptsize
\begin{tabular}
{ccccccc|cccc}
\toprule
&&& \multicolumn{4}{c}{\texttt{GossipCop++}}&\multicolumn{4}{c}{\texttt{PolitiFact++}} \\ \cline{4-11}
&&&\multicolumn{4}{c}{Accurancy w.r.t. each group}&\multicolumn{4}{c}{Accurancy w.r.t. each group} \\ \cline{4-11}
\multirow{2}{*}{{\splitcell{MF portion\\(Training Data)}} 
}&&  &\multicolumn{2}{c}{Real}&\multicolumn{2}{|c}{Fake}&\multicolumn{2}{c}{Real}&\multicolumn{2}{|c}{Fake}  \\ \cline{4-11}& Model size & Model name & HR & MR & HF & MF & HR & MR & HF & MF\\ \hline
\multirow{10}{*}{0\%}&\multirow{5}{*}{Large}&RoBERTa & 29.03 & 94.74 & 92.17 & 4.41 & 16.49 & 91.75 & 84.54 & 4.12 \\
&&BERT & 38.09 & 93.76 & 89.47 & 3.67 & 23.20 & 93.81 & 82.47 & \textbf{7.22} \\
&&ELECTRA & \textbf{39.07} & 95.10 & 86.29 & 10.77 & 12.89 & 94.85 & 81.44 & 2.06 \\
&&ALBERT & 16.35 & 87.64 & \textbf{94.86} & 6.98 & 17.53 & 86.60 & \textbf{91.75} & 6.19 \\
&&DeBERTa & 24.68 & \textbf{96.21} & 93.27 & 7.96 & 13.92 & \textbf{95.88} & 90.72 & 3.09 \\ \cdashline{2-11}
&\multirow{5}{*}{Base}&RoBERTa & 27.62 & 92.66 & 89.11 & 9.67 & 13.40 & 88.66 & 84.54 & 3.09 \\
&&BERT & 29.94 & 91.43 & 85.68 & 6.73 & \textbf{25.77} & 91.75 & 81.44 & 6.19 \\
&&ELECTRA & 34.05 & 93.15 & 84.94 & 3.79 & 22.16 & 92.78 & 86.60 & 1.03 \\
&&ALBERT & 19.41 & 90.45 & 93.02 & 7.96 & 16.49 & 89.69 & 90.72 & 4.12 \\
&&DeBERTa & 17.33 & 91.80 & 94.49 & \textbf{14.20} & 11.34 & 87.63 & 89.69 & 6.19 \\
\hline
\multirow{10}{*}{33\%}&\multirow{5}{*}{Large}&RoBERTa & 18.06 & 89.35 & 95.47 & \textbf{98.04} & 3.09 & 90.72 & 89.69 & \textbf{97.94} \\
&&BERT & 22.11 & 86.41 & 94.49 & 95.72 & 10.31 & 79.38 & 89.69 & 97.94 \\
&&ELECTRA & \textbf{30.25} & \textbf{92.41} & 91.31 & 89.35 & 9.28 & \textbf{91.75} & 90.72 & 91.75 \\
&&ALBERT & 15.74 & 83.72 & 94.12 & 91.80 & 15.46 & 82.47 & 90.72 & 92.78 \\
&&DeBERTa & 18.74 & 91.55 & \textbf{95.72} & 96.21 & 12.89 & 89.69 & \textbf{96.91} & 96.91 \\ \cdashline{2-11}
&\multirow{5}{*}{Base}&RoBERTa & 26.15 & 89.60 & 92.04 & 92.29 & \textbf{18.56} & 83.51 & 82.47 & 93.81 \\
&&BERT & 25.66 & 87.27 & 91.31 & 93.15 & 9.28 & 87.63 & 88.66 & 95.88 \\
&&ELECTRA & 23.03 & 87.76 & 91.31 & 87.03 & 12.89 & 86.60 & 92.78 & 90.72 \\
&&ALBERT & 19.17 & 86.90 & 94.74 & 89.60 & 7.22 & 81.44 & 95.88 & 91.75 \\
&&DeBERTa & 20.58 & 88.74 & 93.27 & 91.06 & 11.34 & 85.57 & 91.75 & 92.78 \\
\hline
\multirow{10}{*}{50\%}&\multirow{5}{*}{Large}&RoBERTa & 23.33 & 89.60 & \textbf{94.00} & \textbf{99.14} & 5.67 & 91.75 & 89.69 & \textbf{100.00} \\
&&BERT & 25.41 & 85.31 & 91.55 & 97.31 & 10.82 & 83.51 & 88.66 &\textbf{100.00} \\
&&ELECTRA & \textbf{32.21} & 91.55 & 90.21 & 94.12 & 13.92 & 91.75 & 86.60 & 95.88 \\
&&ALBERT & 20.70 & 85.43 & 90.33 & 93.64 & 23.20 & 83.51 & 86.60 & 95.88 \\
&&DeBERTa & 27.86 & \textbf{94.00} & 92.41 & 97.67 & \textbf{25.26} & \textbf{92.78} & 89.69 & 98.97 \\ \cdashline{2-11}
&\multirow{5}{*}{Base}&RoBERTa & 29.58 & 88.13 & 90.21 & 94.74 & 22.16 & 81.44 & 83.51 & 95.88 \\
&&BERT & 31.72 & 86.41 & 89.23 & 96.08 & 9.28 & 86.60 & 86.60 & 97.94 \\
&&ELECTRA & 27.80 & 87.15 & 90.58 & 93.51 & 21.65 & 86.60 & 88.66 & 94.85 \\
&&ALBERT & 23.82 & 88.37 & 91.19 & 94.86 & 9.79 & 87.63 & \textbf{92.78} & 97.94 \\
&&DeBERTa & 22.90 & 85.07 & 90.94 & 89.72 & 24.23 & 87.63 & 90.72 & 94.85 \\
\hline
\multirow{10}{*}{67\%}&\multirow{5}{*}{Large}&RoBERTa & 24.49 & 87.39 & \textbf{93.27} & \textbf{99.27} & 11.86 & 87.63 & 88.66 & \textbf{100.00} \\
&&BERT & 34.35 & 84.70 & 89.35 & 97.55 & 12.89 & 83.51 & 81.44 &\textbf{100.00} \\
&&ELECTRA & 39.25 & 91.55 & 85.43 & 97.31 & 24.74 & 90.72 & 80.41 & 96.91 \\
&&ALBERT & 30.92 & 85.56 & 83.11 & 95.59 & 39.18 & 84.54 & 75.26 & 95.88 \\
&&DeBERTa & 30.13 & \textbf{94.49} & 90.70 & 98.78 & 26.29 & \textbf{95.88} & \textbf{90.72} &\textbf{100.00} \\ \cdashline{2-11}
&\multirow{5}{*}{Base}&RoBERTa & 34.29 & 88.86 & 86.78 & 96.94 & 38.66 & 81.44 & 75.26 & 97.94 \\
&&BERT & \textbf{40.54} & 88.00 & 84.82 & 97.18 & 22.16 & 88.66 & 81.44 & 98.97 \\
&&ELECTRA & 33.19 & 86.41 & 89.11 & 96.33 & 39.18 & 82.47 & 82.47 & 95.88 \\
&&ALBERT & 34.97 & 87.76 & 85.92 & 94.61 & 21.65 & 86.60 & 83.51 & 95.88 \\
&&DeBERTa & 28.23 & 84.82 & 88.13 & 93.39 & \textbf{47.94} & 87.63 & 85.57 & 95.88 \\
\hline
\multirow{10}{*}{100\%}&\multirow{5}{*}{Large}&RoBERTa & 85.36 & 85.68 & 43.70 & 99.51 & 89.18 & 88.66 & 36.08 & \textbf{100.00} \\
&&BERT & 90.39 & 90.09 & 26.93 & 98.16 & 69.07 & 89.69 & 28.87 & 98.97 \\
&&ELECTRA & 89.28 & 92.04 & 31.21 & 99.39 & 86.08 & 89.69 & 27.84 &\textbf{100.00} \\
&&ALBERT & \textbf{98.22} & \textbf{97.31} & 5.14 & 95.84 & \textbf{96.39} & \textbf{100.00} & 3.09 & 92.78 \\
&&DeBERTa & 91.79 & 93.76 & 23.99 & 99.51 & 83.51 & 92.78 & \textbf{39.18} & 98.97 \\ \cdashline{2-11}
&\multirow{5}{*}{Base}&RoBERTa & 83.28 & 84.33 & \textbf{46.88} & \textbf{99.63} & 87.11 & 83.51 & 19.59 &\textbf{100.00} \\
&&BERT & 91.18 & 90.94 & 18.36 & 97.92 & 86.08 & 92.78 & 21.65 & 98.97 \\
&&ELECTRA & 84.57 & 89.23 & 39.29 & 97.31 & 84.54 & 89.69 & 34.02 &\textbf{100.00} \\
&&ALBERT & 96.14 & 96.82 & 11.14 & 95.96 & 94.33 & 97.94 & 10.31 & 94.85 \\
&&DeBERTa & 87.32 & 88.98 & 33.17 & 96.70 & 93.81 & 90.72 & 31.96 &\textbf{100.00} \\

\bottomrule
\end{tabular}
\caption{Complete results for the \textit{Machine dominance} setting. }
\label{tab: complete result: All MR}
\end{table*}

\clearpage

\section{Detailed Dataset Analysis} \label{Appendix: dataset}
\begin{figure}[ht]
\begin{subfigure}{.5\textwidth}
  \centering
  \includegraphics[width=.8\linewidth]{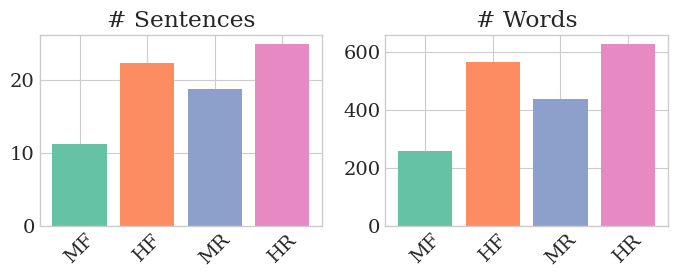}  
  \caption{\texttt{GossipCop++}}
\end{subfigure}
\begin{subfigure}{.5\textwidth}
  \centering
  \includegraphics[width=.8\linewidth]{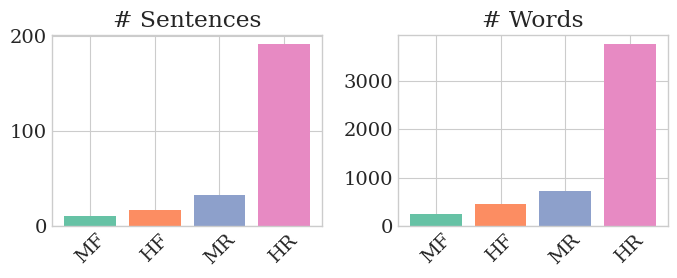}  
  \caption{\texttt{PolitiFact++}}
 
\end{subfigure}

\caption{Average sentence count and average word count density histogram for \texttt{GossipCop++} and \texttt{PolitiFact++}.}
\label{fig: Dataset statistics-average length}
\end{figure}

Figure \ref{fig: Dataset statistics-average length} shows the average sentence count and word count for both \texttt{GossipCop++} and \texttt{PolitiFact++}. We observe that HR generally consists of longer articles compared to other subclasses, while machine-generated news articles tend to be shorter on average, especially MF. Moreover, the figure shows substantial variations in terms of average length across the different datasets.  For instance, when comparing \texttt{GossipCop++} to \texttt{PolitiFact++}, the former has an average of 625 words and 25 sentences, whereas the latter is significantly longer, with 3,759 words and 191 sentences, i.e., seven times larger. Another distinction is that in \texttt{GossipCop++} the average sentence count and word count for HF (22 sentences and  564 words) and HR are quite close to each other. In contrast, within the \texttt{PolitiFact++} dataset, HR is roughly 10 times longer than HF, with HR consisting of 17 sentences and 459 words. Although the total number of news articles in \texttt{PolitiFact++} is too small to train a reliable fake news detector, it serves as a valuable out-of-domain dataset for assessing the robustness of the detector, given its differences from \texttt{GossipCop++}.

\begin{figure}[ht]
\begin{subfigure}{.5\textwidth}
  \centering
  \includegraphics[width=.8\linewidth]{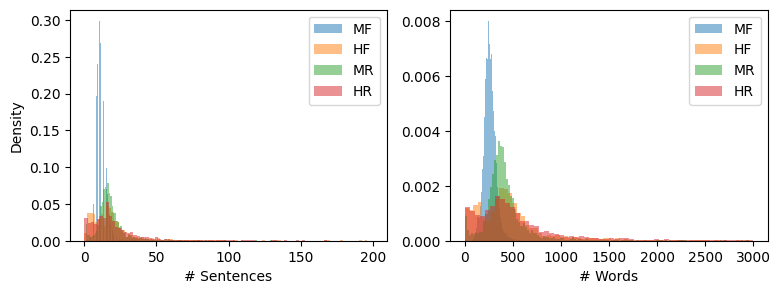}  
  \caption{\texttt{GossipCop++}}
\end{subfigure}
\begin{subfigure}{.5\textwidth}
  \centering
  \includegraphics[width=.8\linewidth]{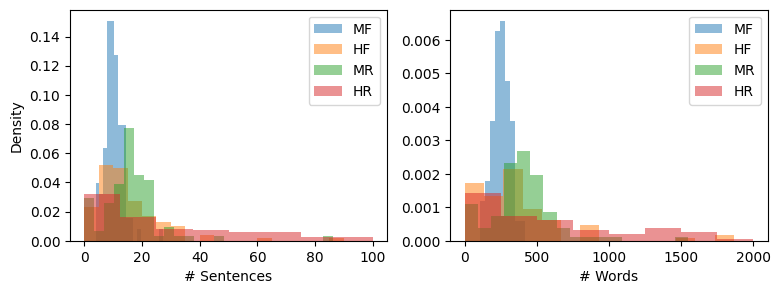}  
  \caption{\texttt{PolitiFact++}}
 
\end{subfigure}

\caption{Sentence count and word count density histogram for \texttt{GossipCop++} and \texttt{PolitiFact++}.}
\label{fig: Dataset statistics}
\end{figure}

In Figure \ref{fig: Dataset statistics}, we randomly extract 4,084 articles in each subclass for \texttt{GossipCop++} and 97 articles in each subclass of \texttt{PolitiFact++} to visualize the distribution of the number of sentences and the number of words for each subclass. Because the HR class in \texttt{PolitiFact++} has extremely long tails, for the ease of representation, we restrict the range of the histogram to be [0;2000] in word count and restrict the $x$ axis to be [0,100] in sentence count. See also Figure~\ref{fig: dataset distribution pairwise comparison-gossipcop} and Figure~\ref{fig: dataset distribution pairwise comparison-politifact} in the Appendix. From Figure~\ref{fig: Dataset statistics}, we find that the distribution of sentence counts and word counts for HF and HR are quite close to each other, spanning a wide range of lengths. Meanwhile, the sentence counts and the word counts for machine-generated articles, especially MF news articles, show more pronounced peaks. 

\subsection{Sentence Length and Word Length}

\begin{figure*}[h]
\begin{subfigure}{.5\textwidth}
  \centering
  \includegraphics[width=.8\linewidth]{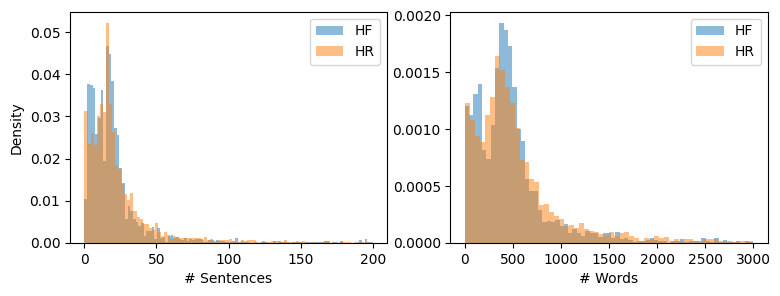}  
  \caption{HF vs. HR}
\end{subfigure}
\begin{subfigure}{.5\textwidth}
  \centering
  \includegraphics[width=.8\linewidth]{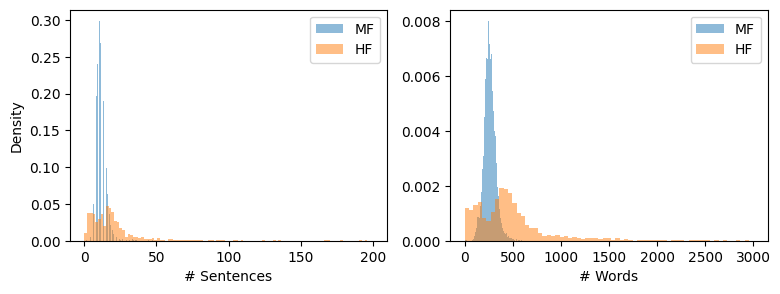}  
  \caption{MF vs. HF}
\end{subfigure}
\begin{subfigure}{.5\textwidth}
  \centering
  \includegraphics[width=.8\linewidth]{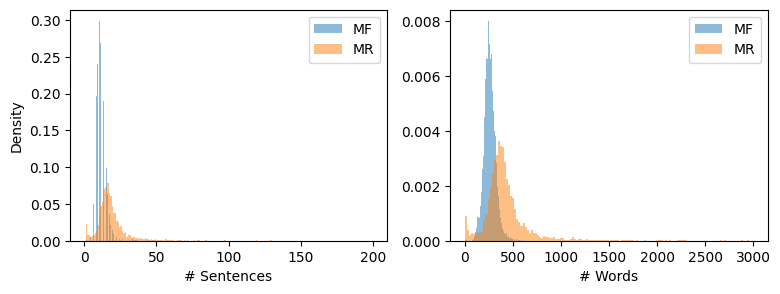}  
  \caption{MF vs. MR}
\end{subfigure}
\begin{subfigure}{.5\textwidth}
  \centering
  \includegraphics[width=.8\linewidth]{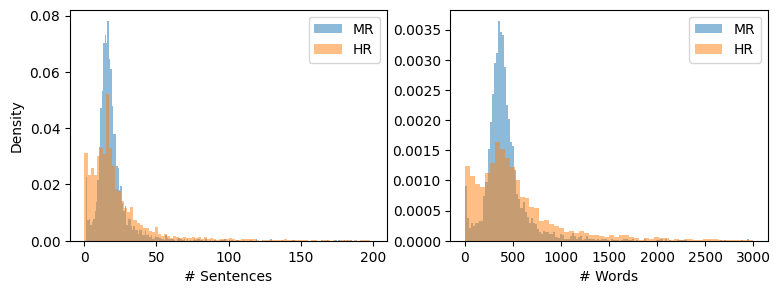}  
  \caption{MR vs. HR}
\end{subfigure}

\begin{subfigure}{.5\textwidth}
  \centering
  \includegraphics[width=.8\linewidth]{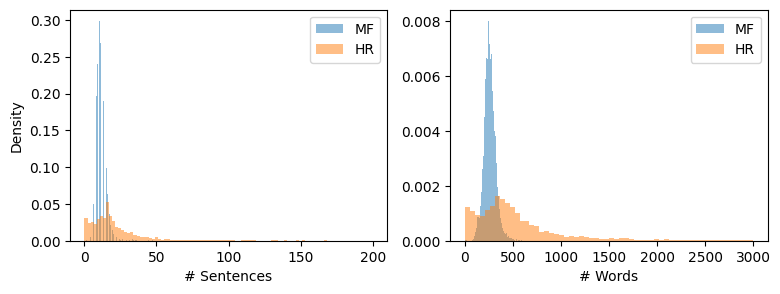}  
  \caption{MF vs. HR}
\end{subfigure}
\begin{subfigure}{.5\textwidth}
  \centering
  \includegraphics[width=.8\linewidth]{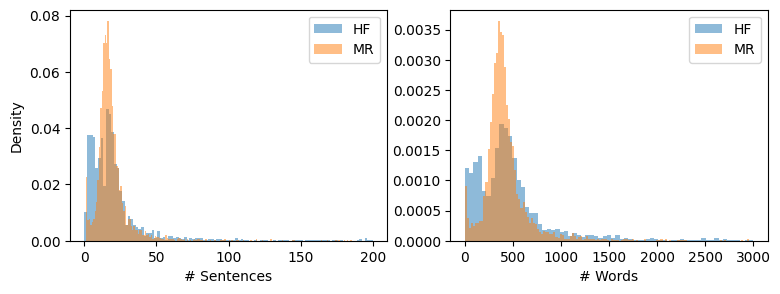}  
  \caption{HF vs. MR}
\end{subfigure}
\caption{Sentence length and word length density histograms for different subclasses in \texttt{GossipCop++}.}
\label{fig: dataset distribution pairwise comparison-gossipcop}
\end{figure*}

\begin{figure*}[h!]
\begin{subfigure}{.5\textwidth}
  \centering
  \includegraphics[width=.8\linewidth]{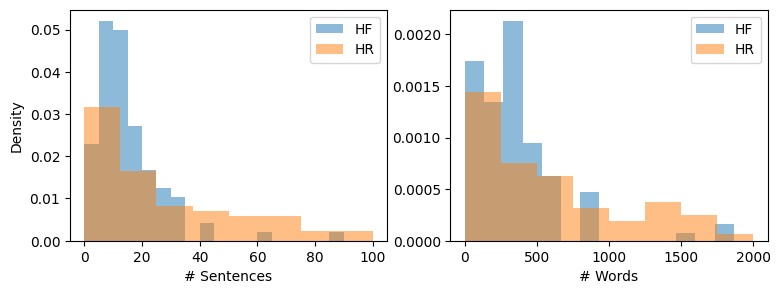}  
  \caption{HF vs. HR}
\end{subfigure}
\begin{subfigure}{.5\textwidth}
  \centering
  \includegraphics[width=.8\linewidth]{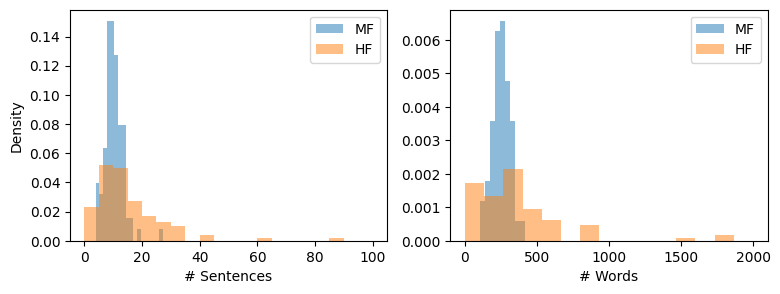}  
  \caption{MF vs. HF}
\end{subfigure}
\begin{subfigure}{.5\textwidth}
  \centering
  \includegraphics[width=.8\linewidth]{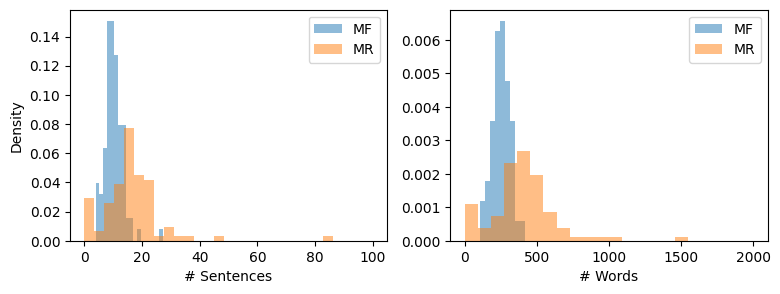}  
  \caption{MF vs. MR}
\end{subfigure}
\begin{subfigure}{.5\textwidth}
  \centering
  \includegraphics[width=.8\linewidth]{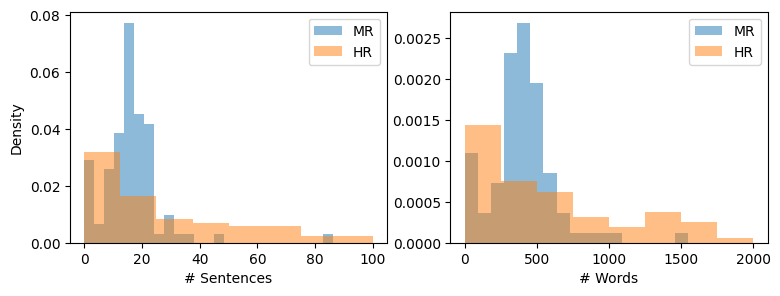}  
  \caption{MR vs. HR}
\end{subfigure}
\begin{subfigure}{.5\textwidth}
  \centering
  \includegraphics[width=.8\linewidth]{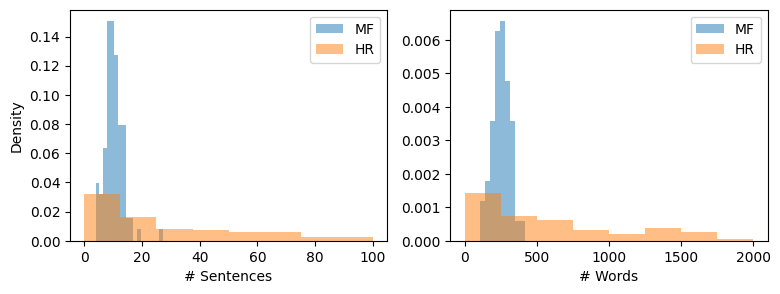}  
  \caption{MF vs. HR}
\end{subfigure}
\begin{subfigure}{.5\textwidth}
  \centering
  \includegraphics[width=.8\linewidth]{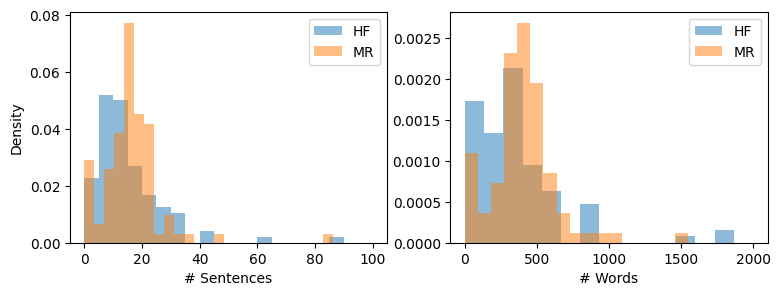}  
  \caption{HF vs. MR}
\end{subfigure}
\caption{Sentence length and word length density histograms for different subclasses in \texttt{PolitiFact++}.}
\label{fig: dataset distribution pairwise comparison-politifact}
\end{figure*}

Figure \ref{fig: dataset distribution pairwise comparison-gossipcop} and Figure \ref{fig: dataset distribution pairwise comparison-politifact} compare the pair-wise distribution of the sentence counts and the word counts. We can see that the distribution of sentence counts and word counts for HF and HR exhibit remarkable similarity. This implies that human-written news articles, regardless of their authenticity, share a significant resemblance in their structural composition. Conversely, there exists a more pronounced disparity in the case of machine-generated news articles (MF and MR), implying that it might be easier to distinguish the veracity of such articles based on their length distribution. Moreover, we observed a notable discrepancy in the distribution of MR and HR subclasses, even though MR is paraphrased from real news articles with approximately the same sentence and word counts.

Although the dataset statistics show a distribution discrepancy between human-written and machine-generated real and fake news, which might be a signal for the current fake news detection problem, 
from a broader data distribution standpoint, if journalists increasingly adopt LLMs in their writing, over time, the distribution of real news articles might gradually shift towards the  distribution of the machine-generated articles (MF and MR). Eventually, this shift could lead to a convergence where the distributions of real and fake news articles once again closely resemble each other.

\clearpage

\section{Comparing Different Detectors in the \textit{Transitional Coexistence} and the \textit{Machine Dominance} Setting.}

Here, we compare different detectors in the \textit{Transitional Coexistence} and the \textit{Machine Dominance} setting as supplementary experiments for Section~\ref{subsec: detectors}.

\subsection{Impact of the Detector Structure}
\begin{figure*}[ht]
\begin{subfigure}{.5\textwidth}
  \centering
  \includegraphics[width=.8\linewidth]{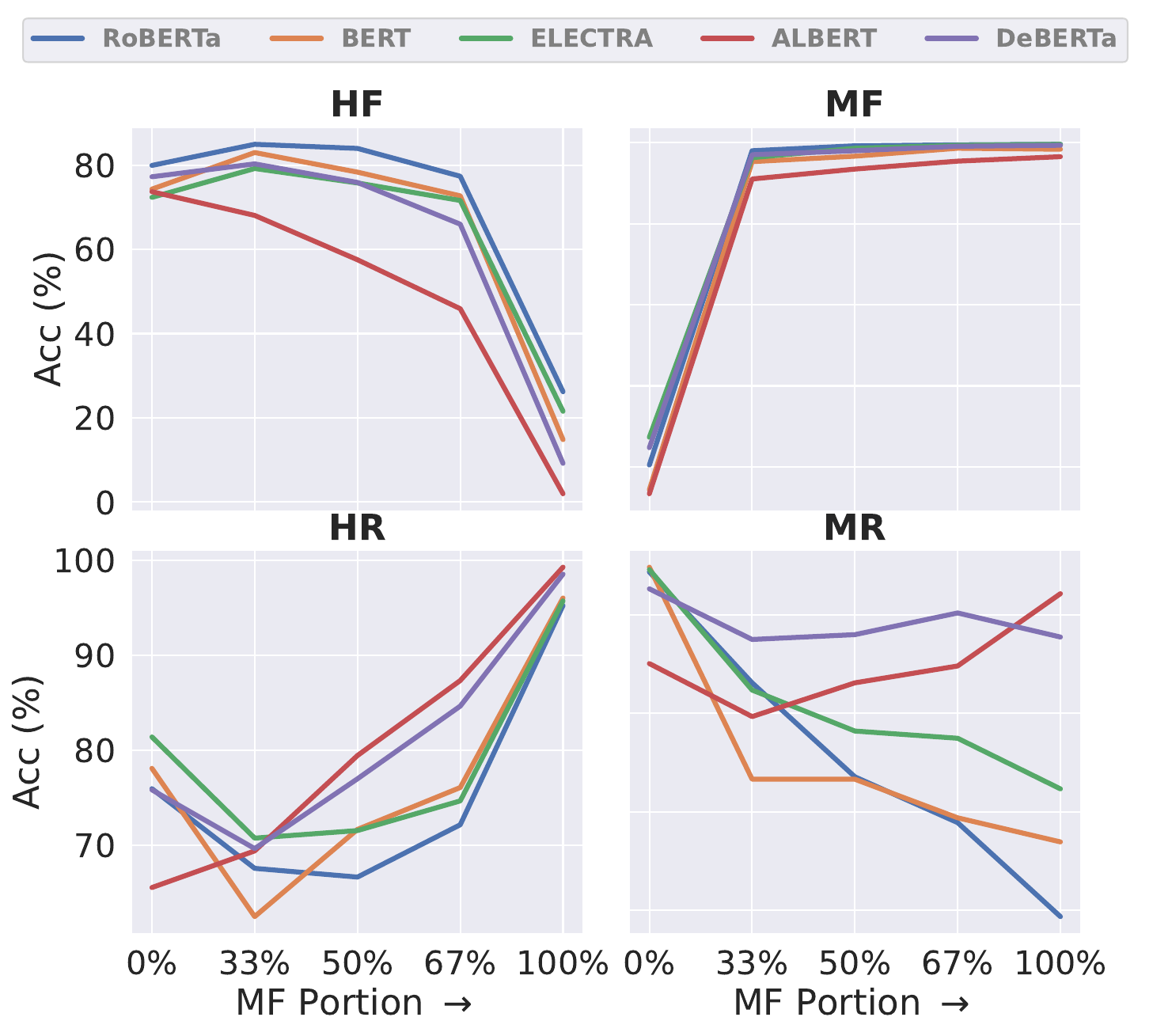}  
  \caption{\textit{Transitional Coexistence}}
\end{subfigure}
\begin{subfigure}{.5\textwidth}
  \centering
  \includegraphics[width=.8\linewidth]{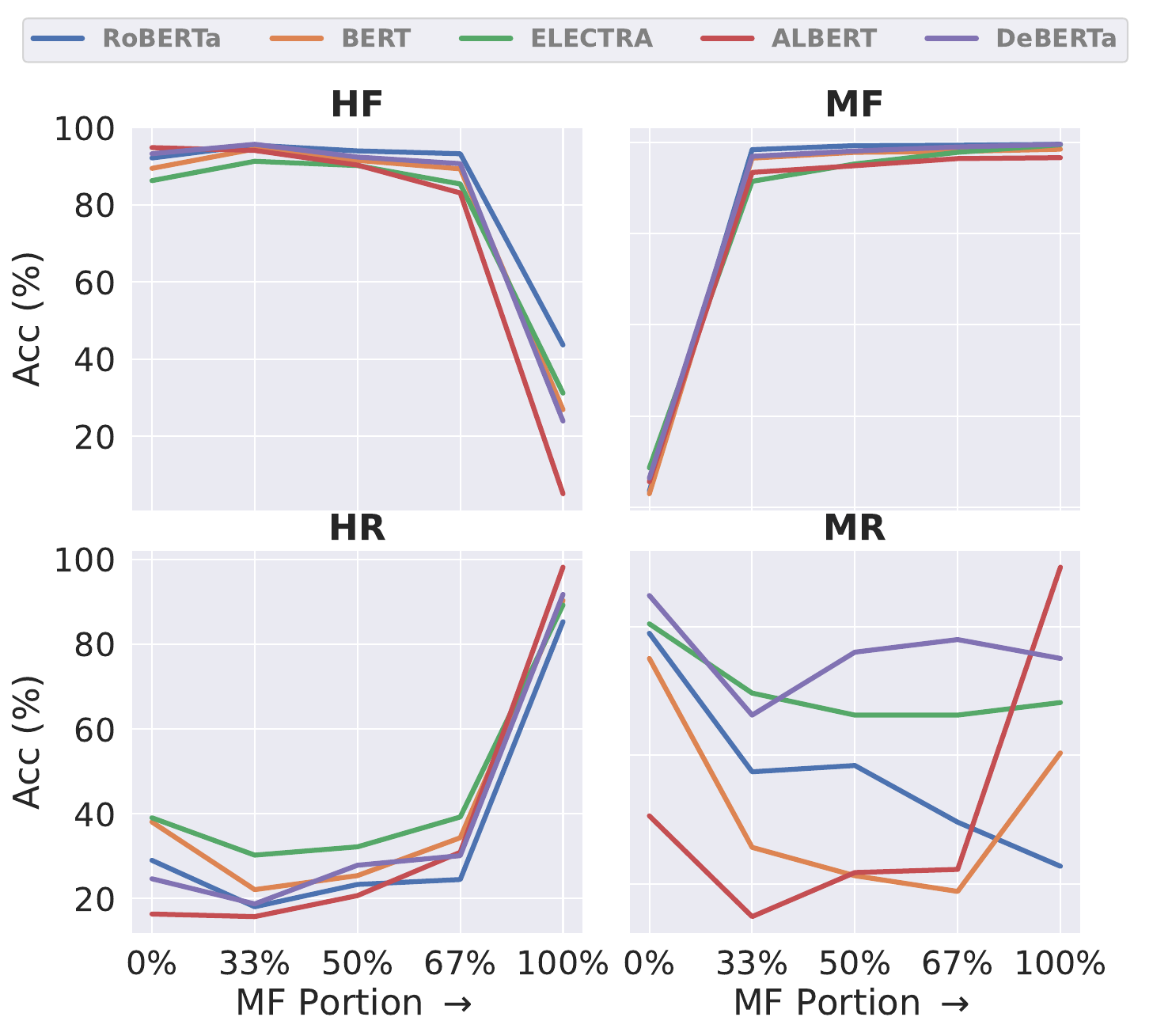}  
  \caption{\textit{Machine Dominance}}
\end{subfigure}
\caption{Comparing different detectors (RoBERTa, BERT, ELECTRA, ALBERT, DeBERTa) in the \textit{Transitional Coexistence} and the \textit{Machine Dominance} settings.}
\label{fig: compare-detectors-MD+Tr}
\end{figure*}

\subsection{Inpact of the Detector Size}
\begin{figure*}[ht]
\begin{subfigure}{.5\textwidth}
  \centering
  \includegraphics[width=.8\linewidth]{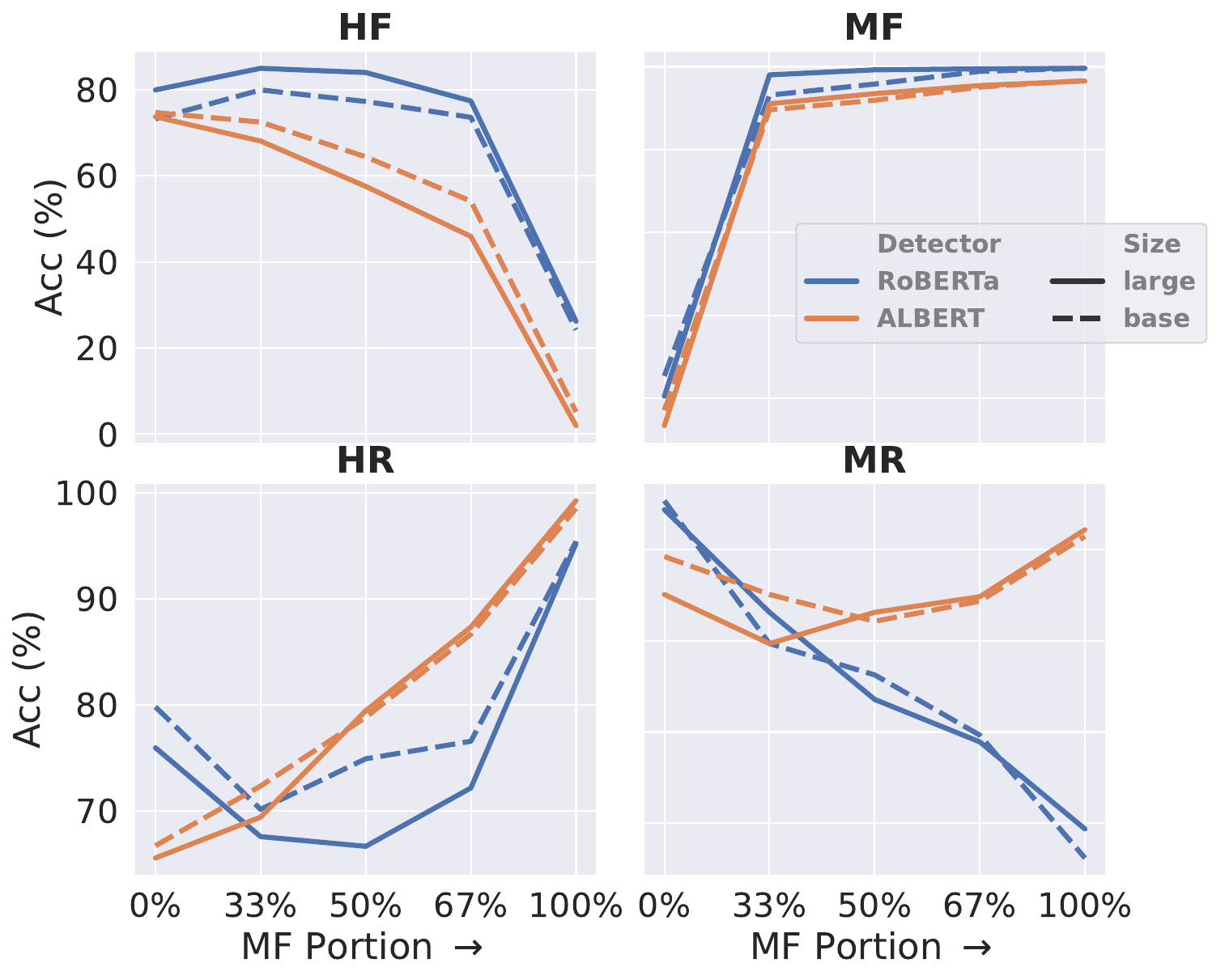}  
  \caption{\textit{Transitional Coexistence}}
\end{subfigure}
\begin{subfigure}{.5\textwidth}
  \centering
  \includegraphics[width=.8\linewidth]{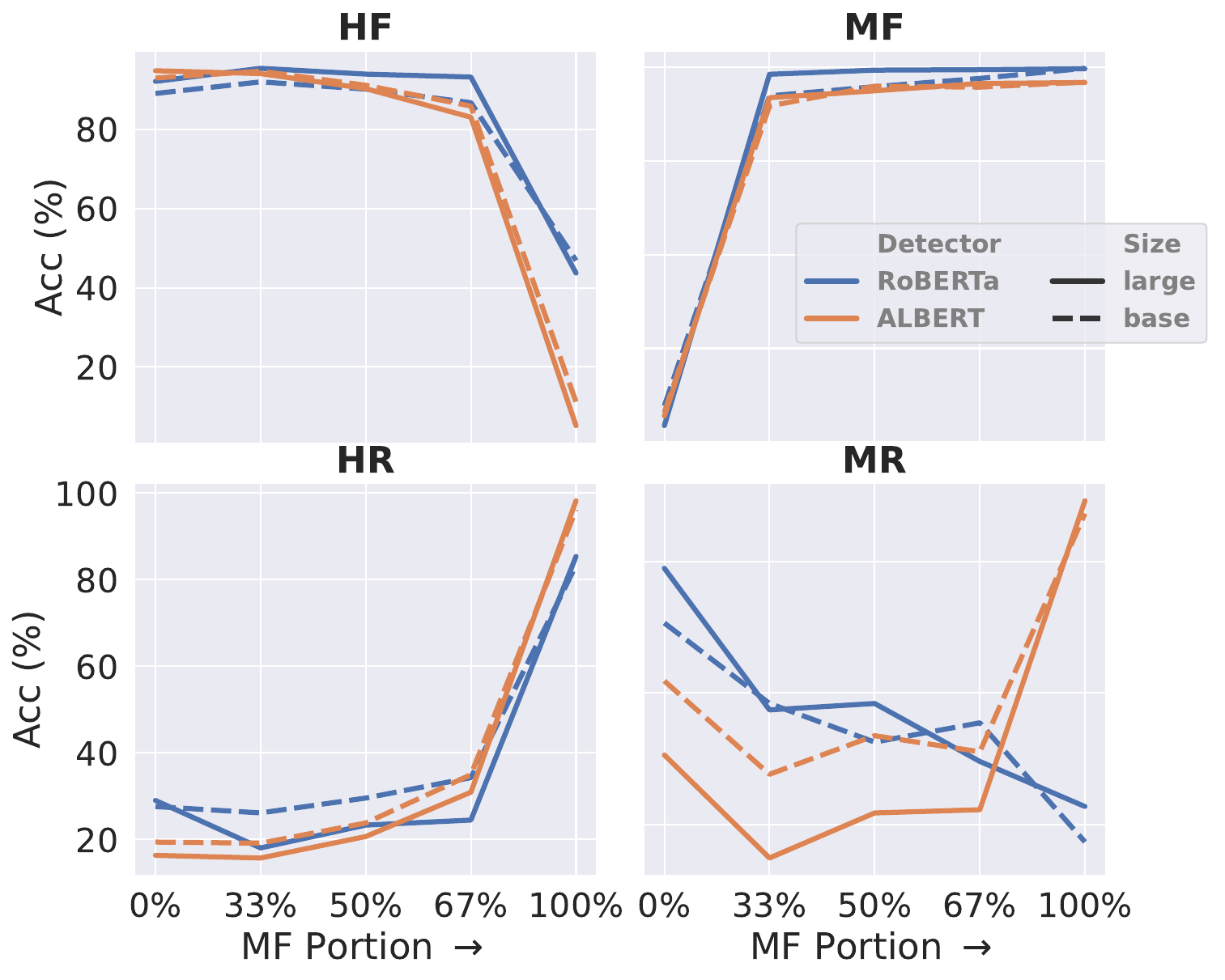}  
  \caption{\textit{Machine Dominance}}
\end{subfigure}
\caption{Comparing RoBERTa and ALBERT detectors in the \textit{Transitional Coexistence} and the \textit{Machine Dominance} settings for models of different sizes: large vs. base models.}
\label{fig: compare-detectors-size-MD+Tr}
\end{figure*}

\end{document}